\pdfoutput=1

\documentclass[11pt]{article}

\usepackage[final]{acl}

\usepackage{times}
\usepackage{latexsym}

\usepackage[T1]{fontenc}

\usepackage[utf8]{inputenc}

\usepackage{microtype}

\usepackage{inconsolata}

\usepackage{graphicx}

\usepackage{booktabs} 
\usepackage{siunitx} 

\usepackage{bbm}
\usepackage{bm}
\usepackage[inline]{enumitem}
\usepackage{amsmath,amssymb}
\usepackage{multirow}

\graphicspath{ {figures/} }

\newcommand{\method}{\texttt{KEEN}}

\newcommand{\nl}[1]{\textit{``#1''}}

\definecolor{mypurple}{RGB}{103,78,167}
\definecolor{myblue}{RGB}{65,124,218}
\definecolor{mygreen}{RGB}{148,193,129}

\title{Estimating Knowledge in Large Language Models \\ Without Generating a Single Token}

\usepackage{authblk}
\author{Daniela Gottesman}
\author{Mor Geva}
\affil{Blavatnik School of Computer Science, Tel Aviv University}
\affil{{\texttt{gottesman3@mail.tau.ac.il}}, {\texttt{morgeva@tauex.tau.ac.il}}}

\begin{document}
\maketitle

\begin{abstract}

To evaluate knowledge in large language models (LLMs), current methods query the model and then evaluate its generated responses. In this work, we ask whether evaluation can be done \textit{before} the model has generated any text. Concretely,
is it possible to estimate how knowledgeable a model is about a certain entity, only from its internal computation?
We study this question with two tasks: given a subject entity, the goal is to predict (a) the ability of the model to answer common questions about the entity, and (b) the factuality of open-ended responses generated by the model about the entity.
Experiments with a variety of LLMs show that \method{},
a simple probe trained over internal subject representations, succeeds at both tasks --- correlating with both the QA accuracy of the model per-subject and FActScore, a recent factuality metric in open-ended generation.
Moreover, \method{} naturally aligns with the model's hedging behavior and faithfully reflects changes in the model's knowledge after fine-tuning. Lastly, we show a more interpretable yet equally performant variant of \method{}, which highlights a small set of tokens indicative of clusters and gaps in the model's knowledge.
Being simple and lightweight, \method{} can be leveraged to guide decisions such as when it is appropriate to apply further training or augment queries with retrieval.

\end{abstract}

\section{Introduction}
 
The standard approach for evaluating knowledge in large language models (LLMs) relies on querying the model, letting it generate responses, and then evaluating the responses. This evaluation can be done using various methods, including comparing responses to gold answers \cite{touvron2023llama, cohen-etal-2023-crawling}, measuring response consistency over multiple generations \citep{cohen-etal-2023-lm, manakul-etal-2023-selfcheckgpt, kuhn2023semantic}, checking the support of responses in external evidence \cite{gao-etal-2023-rarr, bohnet2022attributed}, or estimating the model's uncertainty per-response \cite{yu2024mechanisms, yuksekgonul2024attention, li2023inferencetime, snyder2023early, liu-etal-2022-token}.

In this work, we take a step back and ask whether it is possible to evaluate the model's knowledge \textit{before} it generates any text, using only its internal computation.
This view is analogous to human studies that show the effectiveness of assessing non-verbal communication for determining witness credibility in the courtroom \cite{remland1994importance, denault2024use}.
Concretely, we propose to evaluate how knowledgeable an LLM is about a given subject entity (e.g. Napoleon or Empire State Building), by considering only how it processes the name of that entity, and \textit{before} it generates a single token.

\begin{figure}[t]
    \centering
    \includegraphics[scale=0.33]{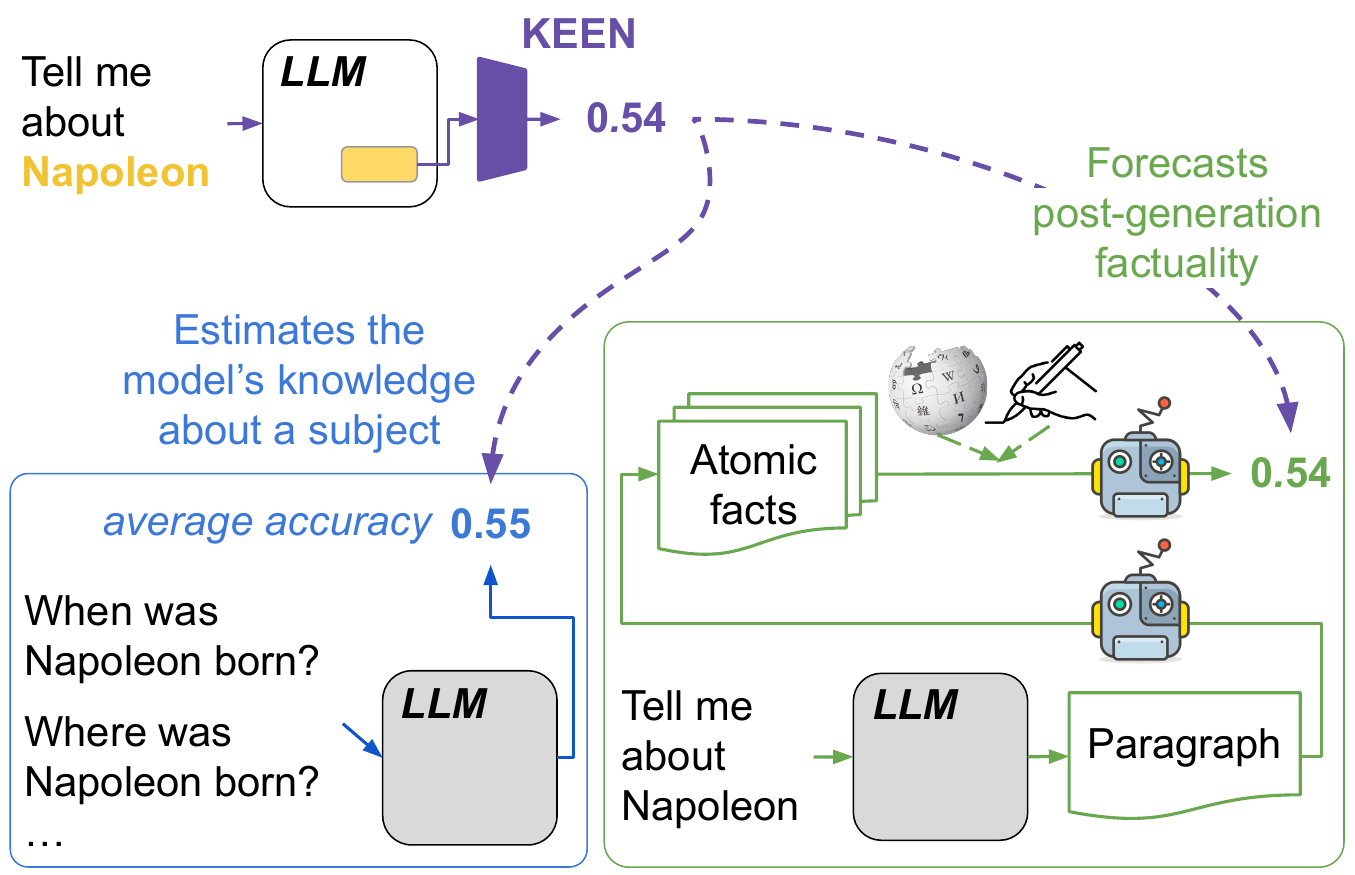}
    \caption{We show that simple probes (\textbf{\textcolor{mypurple}{\method{}}}), trained over hidden model representations, quantify the model's knowledge about a given subject entity --- estimating the model's question-answering accuracy on entity-related questions (\textbf{\textcolor{myblue}{bottom left}}) and forecasting the factuality of model-generated texts about the entity (\textbf{\textcolor{mygreen}{right}}).}
    \label{fig:intro}
\end{figure}

We formalize this problem as entity knowledge estimation (\S\ref{sec:problem_setup}) 
and devise two concrete tasks. Given an entity, the goal is to predict: (a) how many common questions about the subject entity the model will answer correctly (Figure~\ref{fig:intro}, bottom left), and (b) how many of the claims in a model generated response about the subject are factually correct (Figure~\ref{fig:intro}, right).

To tackle entity knowledge estimation, we capitalize on findings from recent interpretability works which show that, during inference, the hidden representations of an input entity capture many attributes related to it \cite{geva-etal-2023-dissecting, meng-locating}, and often these attributes can be extracted with linear functions \cite{hernandez2024linearity}. 
Therefore, we propose (\S\ref{sec:keen}) to estimate how knowledgeable a model is about a given entity by training simple probes, called \method{} (Knowledge Estimation of ENtities), over the model's representations of the entity (Figure~\ref{fig:intro}, upper left).

We evaluate \method{} in two experimental settings (\S\ref{sec:experiments}) of factual question answering (QA) and open-ended generation (OEG) of biographies. 
In the QA setting, we derive a set of questions per-subject for subjects in PopQA \citep{mallen-etal-2023-trust} and evaluate how well \method{} predicts the model's average accuracy per-subject across these questions.
In the OEG setting, we evaluate the correlation of \method{} with FActScore \citep{min-etal-2023-factscore}, a post-generation hallucination detector. 
In both settings and across models of different sizes and families  --- GPT2 \cite{Radford2019LanguageMA}, Pythia \cite{10.5555/3618408.3618510}, LLaMA2 \cite{touvron2023llama}, and Vicuna \cite{vicuna2023} --- \method{} consistently shows correlation values between 0.58-0.68 with QA accuracy and 0.66-0.77 with OEG factuality.
Moreover, \method{} probes trained on entity representations show substantially stronger correlation with both metrics than probes trained on commonly-used intrinsic features, such as fully-connected scores and self-attention activations, and external features, such as entity-popularity.

Further analyzing the utility and features of \method{} (\S\ref{sec:analysis}), we show that \method{} faithfully correlates with the model's hedging behavior, i.e., the score predicted by \method{} decreases as the fraction of per-entity questions that a model hedges on increases.
In addition, \method{} faithfully reflects changes in the model's knowledge following fine-tuning: training LLaMA2 on Wikipedia articles about certain entities increases their \method{} score while scores for other entities tend to decrease.
Lastly, we show that training \method{} on the vocabulary projections of entity representations \cite{nostalgebraist2020, geva-etal-2021-transformer} increases the probe's interpretability without performance cost,
identifying a small set of tokens that represent clusters or gaps in entity knowledge.

To conclude, we present \method{}, a simple and lightweight approach for quantifying how knowledgeable a model is about a given entity from intrinsic properties, which well-estimates the accuracy and factuality of model responses about the entity.
We also show that \method{} scores are reflective of both hedging behavior and changes in entity-based knowledge after fine-tuning. 
Practically, \method{} could be used to inform developer decisions such as whether to augment queries with retrieval, discard certain queries (e.g. by abstaining), enhance models with external tools, or identify ``holes'' in the model's knowledge to apply further training on.
We release our code and data at \url{https://github.com/dhgottesman/keen_estimating_knowledge_in_llms}.

\section{Entity Knowledge Estimation}
\label{sec:problem_setup}

Our goal is to evaluate how much knowledge an LLM captures about an entity from how it processes the entity's name alone, without obtaining model responses and evaluating them post-generation. 
This view is motivated by growing evidence from interpretability works which find that, during model inference, knowledge is centralized in the hidden representations corresponding to named entities \citep{meng-locating, geva-etal-2023-dissecting, li-etal-2021-implicit}.

Given a subject entity $s$ (e.g. Napoleon or Empire State Building) and a model $M$, our goal is to estimate two related quantities: (a) the performance of $M$ on queries about $s$, and (b) the probability that $M$ will generate incorrect facts given any query about $s$.
These two quantities are expected to be related, as they are both influenced by and reflect the amount of knowledge $M$ captures about $s$.

To evaluate entity knowledge, we propose two concrete evaluation settings:

\paragraph{Question Answering (QA)} For a subject entity $s$ and a set of common question-answer pairs $\mathcal{Q} = \{\langle q_i, a_i \rangle\}_{i=1}^n$ about $s$, denote by $\hat{a}_i$ the answer predicted by a model $M$ for the query $q_i$. Given only the subject $s$, our goal is to estimate the average accuracy of $M$ over $\mathcal{Q}$, denoted as ${y^{(s)}_{QA} := \frac{1}{n}\sum_{i=1}^n \mathbbm{1}[\hat{a}_i = a_i]}$. 

\paragraph{Open-Ended Generation (OEG)}
For a general information-seeking query $q$ about a subject $s$ (e.g. \nl{Tell me facts about Napoleon} or \nl{Generate a paragraph about Napoleon}), let $\mathcal{R} = \{ \langle c_i, a_i \rangle \}_{i=1}^m$ be the set of claims in the response generated by $M$, each with a 0/1 label indicating its correctness with respect to external evidence ($c_i$ denotes a claim and $a_i$ is its factuality label). 
Claims can be extracted and evaluated for correctness using various automatic methods \citep[e.g.,][]{nenkova-passonneau-2004-evaluating, shapira-etal-2019-crowdsourcing, zhang-bansal-2021-finding}. Given only the subject $s$, the task is to predict the portion of factually correct claims in $\mathcal{R}$, denoted as ${y^{(s)}_{OEG} := \frac{1}{m} \sum_{i=1}^m a_i}$.

\vspace{5px}
\noindent A naive solution for both tasks would be to first obtain queries about $s$, feed them to $M$, and evaluate the answers $M$ generates. Here we seek an efficient solution, which estimates the knowledge of $M$ about $s$, without iteratively executing $M$.

\section{\method{}}
\label{sec:keen}

\citet{geva-etal-2023-dissecting} showed that for a given subject in the input, LLMs construct an information-rich representation of the subject that encodes many of its attributes. Furthermore, subject attributes can be extracted from the subject representation with a simple linear function \citep{hernandez2024linearity}.
We capitalize on these findings and propose to train a simple probe over the model's representations of subjects to predict how much knowledge the model captures about them. In our following formulation (and the rest of the paper), we focus on widely-adopted transformer-based auto-regressive language models.

\paragraph{Notation}
Assuming a language model with $L$ layers, a hidden dimension $d$, a vocabulary $\mathcal{V}$, and an unembedding matrix $W_U \in \mathbb{R}^{|\mathcal{V}| \times d}$.
Let $\mathbf{h}_{\ell,i}$ be the hidden representation at position $i$ and layer $\ell$, omitting normalization, $\mathbf{h}_{\ell,i}$ is computed as:
$$\mathbf{h}_{\ell,i} = \mathbf{h}_{\ell-1,i} + \mathbf{a}_{\ell,i} + \mathbf{m}_{\ell,i}$$
where $\mathbf{a}_{\ell,i}$ and $\mathbf{m}_{\ell,i}$ denote the outputs from the $\ell$-th multi-head self-attention and MLP sublayers, respectively \citep{NIPS2017_3f5ee243}.

\subsection{Features}
\label{subsec:features}
Let $t^{(s)}_{1}, ..., t^{(s)}_{s_r}$ be the sequence of $s_r$ input tokens corresponding to a given subject $s$ (e.g. \texttt{N, ap, oleon} for the subject Napoleon tokenized with GPT2).
We use the representations at the last subject position ($s_r$), denoted as $\mathbf{h}^{(s)}_{1,s_r}, ..., \mathbf{h}^{(s)}_{L,s_r}$, to construct a feature vector $\mathbf{z}^{(s)} \in \mathbb{R}^{d_{z}}$.\footnote{
In practice, we obtain the hidden representations using the query: \texttt{``This document describes [$s$]''}. This is to avoid placing the subject in the first position of the input, which often encodes biases that could affect performance on our task \citep{xiao2024efficient, geva-etal-2023-dissecting}. 
}

We train different variants of \method{} probes, each taking as input one of the following sets of features for $\mathbf{z}^{(s)}$:
\begin{itemize}
[itemsep=1pt, topsep=2pt,leftmargin=*]
    \item \textbf{Hidden states (HS)}: We take the subject representation from multiple upper-intermediate layers, where attributes of the subject are often extracted during inference \cite{geva-etal-2023-dissecting, meng-locating} and are easier to disentangle \cite{huang2024ravel, hernandez2024linearity}. 
    To account for variations in the inference pass of different subjects, we choose 3 consecutive layers $\mathcal{L}= \left\{ \frac{3}{4}L + k \mid k \in \{-1, 0, 1\} \right\}$, from which we extract the hidden states ${\{\mathbf{h}^{(s)}_{\ell,s_r} \mid \ell \in \mathcal{L}\}}$. Then, we normalize these vectors (see details below) and average them into a $d$-dimensional feature vector. A systematic evaluation supporting our choice of layers is presented in \S\ref{subsec:layer_config}. 

    \item \textbf{HS with vocabulary projection (VP)}: We take the same hidden states as in HS, but instead of using them as-is, we use their projections to the vocabulary \cite{nostalgebraist2020, geva-etal-2021-transformer}.
    Namely, we normalize and average the vectors ${\{W_U f_L(\mathbf{h}^{(s)}_{\ell, s_r}) \mid \ell \in \mathcal{L}\}}$
    into a $|\mathcal{V}|$-dimensional feature vector, 
    where $f_L$ is the layer norm applied at the last layer of the model.
    While VP is not expected to improve performance, it could enhance interpretability, as the learned weight for each token signifies feature importance in quantifying subject-related knowledge.

    \item \textbf{HS with top-$k$ of vocabulary projection (VP-$k$)}: 
    Since the vocabulary space is typically large, in order to make the probe more interpretable and efficient, we perform feature selection over the trained VP probe to extract the $k$ most influential tokens from the vocabulary projections. We then normalize and average the obtained $3*k$ features ($k$ for each layer) to train a new smaller probe over $k$-dimensional feature vectors.

\end{itemize}

For each of HS, VP, and VP-$k$, we apply Min-Max normalization before averaging the extracted vectors, which scales each feature to be within $[0, 1]$. For example, after extracting the hidden states $\{\mathbf{h}^{(s)}_{\ell, s_r} \mid \ell \in \mathcal{L}\}$ for some subject $s$, 
we normalize the values of every entry $i\in [d]$ and layer $\ell\in \mathcal{L}$ over a set of subjects $\mathcal{S}$. 
Let $\hat{\mathbf{h}}_{\ell,s_r}^{(s)} \in \mathbb{R}^d$ be the normalized $\mathbf{h}_{\ell,s_r}^{(s)}$, so the feature vector for HS is defined as $\mathbf{z}^{(s)} = \frac{1}{|\mathcal{L}|} \sum_{\ell \in \mathcal{L}} \hat{\mathbf{h}}_{\ell,s_r}^{(s)} \in \mathbb{R}^d$.

\subsection{Probing}

We define the following probe for predicting the model's QA accuracy $y^{(s)}_{QA}$ or response factuality $y^{(s)}_{OEG}$ given the features $\mathbf{z}^{(s)}$ for a subject $s$: 
\begin{equation}\label{eq:probe}
    f(\mathbf{z}) := \sigma (\bm{\theta} \cdot \mathbf{z})
\end{equation}
Where $\sigma$ is the sigmoid function and $\bm{\theta}\in \mathbb{R}^{d_{z}}$ is a single linear transformation. The sigmoid non-linearity is necessary to aid the model in learning scores in the range $[0,1]$.\footnote{We also experimented with linear probes and found that they tended to converge to scores in a narrow range around 0.5, failing to capture the signals in the inputs.}

For each of the two tasks $T\in \{\text{QA, OEG}\}$, we optimize $\bm{\theta}$ over features and labels collected for a set of subjects $\mathcal{S}$ by minimizing the MSE loss:
$$ \mathcal{L}_{MSE}(\bm{\theta}) = \| y^{(s)}_{T} - \sigma (\bm{\theta}\cdot \mathbf{z}^{(s)}) \|_2^2 $$
For more details on the probes' training, see \S\ref{sec:training_details}.

\begin{table*}[t]
\footnotesize
\centering
\begin{tabular}{p{1.6cm}ccl}
\toprule
Input & Task & Output  & Example $ \langle \text{question, model answer} \rangle$ / $ \langle \text{claim, correctness label} \rangle$ pairs \\
subject $s$ &  &  $y^{(s)}_{QA} \;/\; y^{(s)}_{OEG}$ & from $\mathcal{Q} \;/\; \mathcal{R}$  \\
\midrule
\multirow{6.5}{=}{George Washington} & \multirow{3}{*}{QA} & & $\langle \text{In what city was George Washington born?, Westmoreland County} \rangle,$\\
 &  & 0.67 & $\langle \text{What is the religion of George Washington?, Episcopal Church} \rangle$ \\
& & & $\langle \text{Who is the father of George Washington?, Augustine Washington} \rangle$ \\
\cmidrule{2-4}
 & \multirow{3}{*}{OEG} &  & $\langle \text{George Washington was a military man.}, 1 \rangle,$ \\
& & 0.74 & $\langle \text{George Washington was the first President of the United States.}, 1 \rangle,$ \\
& {} & & $\langle \text{He was educated at the College of William and Mary.}, 0 \rangle$ \\
\bottomrule
\end{tabular}
\caption{Example input subject and the expected outputs for the two tasks for Pythia 12B. The output labels were computed based on the average QA accuracy over 12 questions (0.67), and the FActScore score for 35 claims (0.74).}
\label{tab:examples}
\end{table*}

\section{Experiments}
\label{sec:experiments}

In this section, we evaluate \method{} and baselines that rely on different intrinsic and external features. We observed that the VP-50 probe obtained comparable performance while being significantly more interpretable (discussed in \S\ref{subsec:diminish_returns}) so we focus on evaluating the VP and VP-50 variants of this probe.

\subsection{Experimental Setting}
\label{subsec:experimental_setting}

\paragraph{Data}
For the QA task, we sample 3,472 subject entities from PopQA \citep{mallen-etal-2023-trust} and generate a set of 5.3 questions on average per subject. To generate questions, we take subject-relation-object triplets from Wikidata \citep{vrandevcic2014wikidata} and convert them into question-answer pairs with handwritten templates. For instance, the triplet $(\text{Napoleon, place of birth, France})$ will be converted to the question \nl{Where was Napoleon born?} and the answer \nl{France}. In addition, we augment each such example with multiple variants that cover different answer granularities \cite{yona2024narrowing}, accounting for both answer and subject aliases, and handling cases with multiple answers.
We consider a model's prediction for a given subject-relation pair as correct if it contains an exact match with any answer alias in at least one question variation.

For the OEG setting, we use the FActScore dataset \cite{min-etal-2023-factscore}, which includes model-generated biographies, extracted claims, and claim labels which indicate whether the claim is supported or not-supported by the subject's Wikipedia page. We compare our results to the FActScore scores of the same generating model. 

Examples for the two tasks are shown in Table~\ref{tab:examples}.
For both settings, we randomly split each dataset into disjoint sets of subjects: 65\% train, 15\% development, and 20\% test. Importantly, the FActScore dataset and QA train set have a negligible number of overlapping subjects, 1 (0.2\%), which allows us to test transfer learning between the two settings. In \S\ref{sec:dataset_details}, we include additional details regarding dataset generation.

\paragraph{Baselines}
We evaluate three baselines that utilize intrinsic features and external features. 
For intrinsic features, we take the two best variants reported by \citet{snyder2023early}, which trained binary hallucination detectors for QA. 
These detectors use the outputs from the self-attention and MLP modules as features, which were also considered by other recent methods for similar tasks \cite{yu2024mechanisms, yuksekgonul2024attention, li2023inferencetime}.

\begin{itemize}
[itemsep=1pt, topsep=2pt,leftmargin=*]
    \item \textbf{Entity popularity (Pop.)}: It has been established that LLM performance is influenced by entity popularity \cite{mallen-etal-2023-trust, kandpal2023large, yona2024narrowing}. We follow previous works \citep[e.g.,][]{chen-etal-2021-evaluating, mallen-etal-2023-trust, cohen2024evaluating} and approximate entity popularity using statistics from Wikipedia. Concretely, we use the total number of monthly views of the entity's page between the years 2000-2023. 
    \footnote{We also computed thresholded log-popularity as implied by Figure~1 in \citet{mallen-etal-2023-trust}. \method{} is superior across all settings and models, except Vicuna 13B in the OEG setting where correlation increased to 0.65 while \method{} achieved 0.66.}
    
    \item \textbf{Self-attention outputs (ATTN)}: We train the same probe of \method{} (Eq.~\ref{eq:probe}), while using $\mathbf{a}^{(s)}_{L,s_r}$ as the feature vector $\mathbf{z}^{(s)}$, i.e., the output of the last self-attention sublayer for the last input token (which is the last subject token in our setup). 
    
    \item \textbf{Fully-connected activations (FC)}: Here we train a similar probe to ATTN, which sets $\mathbf{z}^{(s)}$ to $\mathbf{m}^{(s)}_{L,s_r}$, the output of the last MLP sublayer for the last input token.
\end{itemize}

\paragraph{Models} We analyze 7 auto-regressive language models across various sizes, latent spaces, and training objectives: GPT2-XL \citep{Radford2019LanguageMA}, Pythia 6B and 12B \cite{10.5555/3618408.3618510}, LLaMA2 7B and 13B \citep{touvron2023llama}, and Vicuna 13B \citep{vicuna2023}.\footnote{Vicuna 7B was also analyzed, but due to its poor accuracy in the QA setting and inconsistent behavior, we omitted it.} The vocabulary sizes range between 30K-50K tokens and the hidden state dimensions range from 4096-5120.

\paragraph{Evaluation}
For every model and subject $s$ in our data, we feed the model a generic prompt \texttt{``This document describes [$s$]''} and extract the features used for all methods: \method{} and the above baselines.
Using these features, we obtain predictions for our two tasks for every method. For the Pop. baseline, we simply take the corresponding popularity value of the subject. 
We report Pearson correlation, associated $p$-values ($p$), and the MSE between the predicted and gold scores, for every task, model and method.
Correlation results are provided in \S\ref{subsec:results}, and the $p$-values and MSE results are reported in \S\ref{sec:additional_results}.

\begin{table}[t]
\setlength\tabcolsep{3.5pt}
\footnotesize
\centering
\resizebox{\linewidth}{!}{
\begin{tabular}{lccccccc} 
\toprule
    & {GPT2} & {Pythia} & {Pythia} & {LLaMA2} & {LLaMA2} & {Vicuna} \\ 
    & {XL} & {6B} & {12B} & {7B} & {13B} & {13B} \\ 
    \midrule
    {Pop.} & {0.30} & {0.32} & {0.28} & {0.27} & {0.25} & {0.26}\\
    {FC} & {0.49} & {0.59} & {0.55} & {0.50} & {0.49} & {0.49} \\
    {ATTN} & {0.53} & {0.63} & {0.60} & {0.58} & {0.50} & {0.52} \\
    VP-50 & {0.54} & {0.64} & {0.59} & {0.53} & {0.48} & {0.50} \\
    VP & \textbf{0.61} & \textbf{0.68} & \textbf{0.64} & \textbf{0.64} & \textbf{0.58} & \textbf{0.60} \\
    {HS} & \textbf{0.60} & \textbf{0.68} & \textbf{0.64} & \textbf{0.64} & \textbf{0.58} & \textbf{0.60} \\
    \bottomrule 
\end{tabular}
}
\caption{Correlation with average QA accuracy for \method{} QA probes and baselines on the QA test set.}
\label{tab:qa_accuracy_corr}
\end{table}

\subsection{Results}
\label{subsec:results}
\paragraph{\method{} well-estimates the model's knowledge about the subject entity}
Tables~\ref{tab:qa_accuracy_corr} and \ref{tab:factscore_corr_oeg} show QA and OEG results, respectively. 

In both settings and across all models, \method{} probes trained on hidden representations and vocabulary projections demonstrate the strongest correlation of 0.60-0.68 with QA accuracy ($p \leq 3.43e^{-70}$) and 0.66-0.77 with FActScore ($p \leq 4.02e^{-6}$). This shows that it is possible to predict how knowledgeable a model is about an entity from the entity's hidden representations.

Predicting factuality based on common intrinsic features (FC and ATTN) consistently under-performs with respect to \method{}, further supporting the finding that entity knowledge is centralized in entity representations during inference.
Furthermore, the entity popularity baseline (Pop.) performs poorly on both tasks, with low correlation values of $\leq 0.32$ in QA and $\leq 0.36$ in OEG. This shows that while external statistics of popularity (such as Wikipedia page count) are useful in deriving general performance trends, they often fail to provide fine-grained entity-level predictions.   

Surprisingly, for the Pythia models even the \method{} OEG VP-50 probe strongly correlates \citep{corr_article, corr_article2} with FActScore, indicating that there is a relatively small set of tokens which are influential in increasing and decreasing predicted accuracy. We further analyze these tokens in \S\ref{subsec:feature_analysis} and provide intuition for interpreting them. Moreover, we discuss the trade-off between interpretability and score correlation in \S\ref{subsec:diminish_returns}.

\begin{table}[t]
\centering
\setlength\tabcolsep{4.5pt}
\footnotesize
\begin{tabular}{lcccccc}
\toprule
    Model & Pop. & {FC} & {ATTN} & {VP-50} & {VP} & {HS} \\
    \midrule 
    Pythia 12B  & {0.36} & {0.61} & \textbf{0.77} & {0.72} & {0.75} & \textbf{0.77} \\ 
    Vicuna 13B  & {0.37} & {0.49} & {0.65} & {0.55} & \textbf{0.66} & \textbf{0.66} \\ \bottomrule
\end{tabular}
\caption{Correlation with FActScore for \method{} OEG probes and baselines on the OEG test set.}
\label{tab:factscore_corr_oeg}
\end{table}

\begin{table}[t]
\centering
\setlength\tabcolsep{4.5pt}
\footnotesize
\begin{tabular}{lcccccc}
\toprule
    Model & Pop. & {FC} & {ATTN} & {VP-50} & {VP} & {HS} \\
    \midrule 
    Pythia 12B & {0.47} & {0.41} & {0.55} & {0.40} & {0.57} & \textbf{0.60} \\
    Vicuna 13B & {0.40} & {0.52} & {0.50} & {0.48} & {0.61} & \textbf{0.62} \\ \bottomrule
\end{tabular}
\caption{Transfer learning results, showing the correlation between FActScore and \method{} QA probes and baselines. Results are reported over the full OEG dataset i.e. all 500 subjects in the unlabeled FActScore dataset.}
\label{tab:factscore_corr_qa}
\end{table}

\paragraph{\method{} QA probes generalize to predict factuality in OEG}
Since knowledge is centralized in the internal representations of entities, their use in estimating knowledge should transfer across different settings. Table~\ref{tab:factscore_corr_qa} shows that the predictions of \method{} QA probes have a moderate to strong correlation \citep{corr_article, corr_article2} of 0.60-0.62 with FActScore ($p \leq 2.12e^{-5}$). Further, the correlation of \method{} QA probes with QA accuracy and FActScore are notably similar, e.g. 0.60 and 0.62 for Vicuna 13B \method{} QA HS probes, respectively. These results show that HS and VP features capture signals that generalize across settings, regardless of whether the task requires explicit (QA) or implicit (OEG) recall of factual knowledge by the model.

\section{Analysis}
\label{sec:analysis}

In this section, we further look into the predictions and features of \method{}, evaluating its faithfulness with respect to model hedging (\S\ref{subsec:hedging}) and changes in the model's knowledge following training (\S\ref{subsec:training}). In addition, we analyze its errors (\S\ref{subsec:error_analysis}) and the features of its VP-50 variant (\S\ref{subsec:feature_analysis}).

\subsection{Correlation with Model Hedging}
\label{subsec:hedging}

To prevent factually incorrect responses, LLMs are trained to hedge in cases of uncertainty, for example by generating \texttt{``I don't know''} \cite{ganguli2023capacity}. Therefore, it is expected that models generally hedge on entities they are less knowledgeable about.
Since the \method{} QA probe score estimates entity-based knowledge, we hypothesize that it should correlate with the fraction of questions that a model hedges on about the entity. 

\begin{figure}[t]
\centering
\includegraphics[scale=0.248]{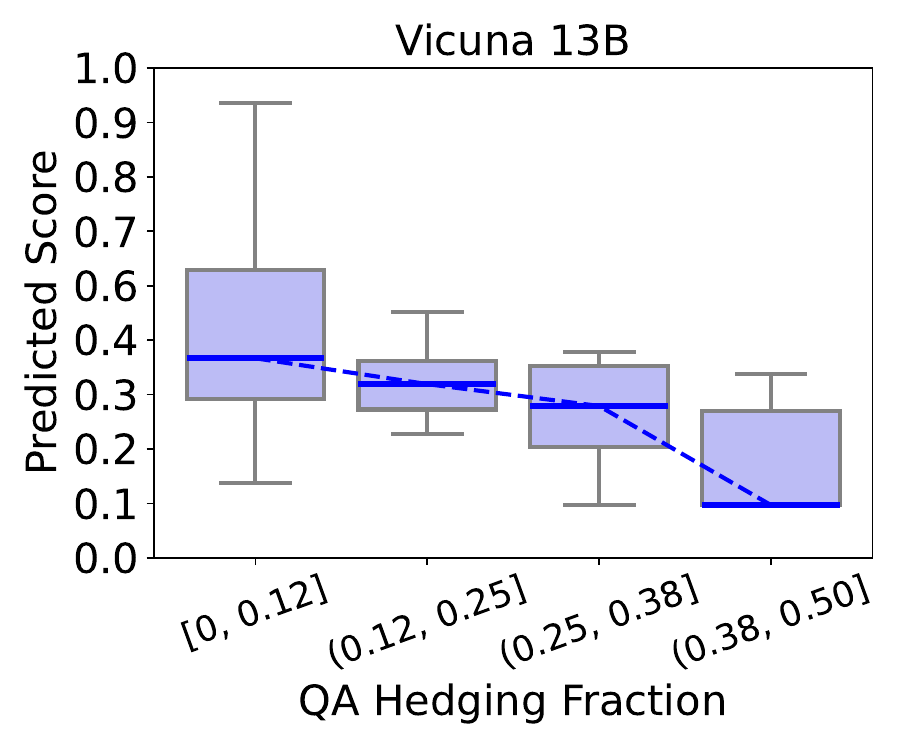}
\includegraphics[scale=0.248]{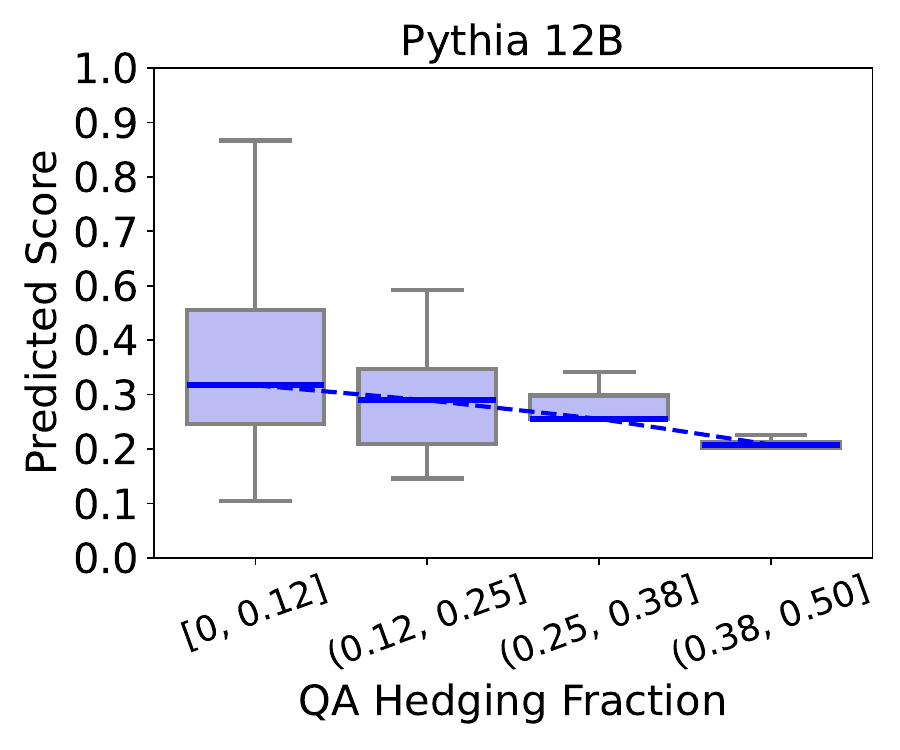}
\caption{\method{} QA scores as a function of the fraction of per-subject queries that Vicuna 13B and Pythia 12B hedge on.}    
\label{fig:vicuna_hedging}
\end{figure}

Figure~\ref{fig:vicuna_hedging} confirms this hypothesis, showing that the \method{} QA VP score decreases as the fraction of queries the model hedges on increases.
This implies that models may hedge based on features of the model's internal representations of the entity, similarly to \method{}. Further details regarding the choice of hedging phrases are provided in \S\ref{subsec:model_hedging_details}.

\subsection{Reflecting Changes in Model Knowledge}
\label{subsec:training}

\begin{figure}[t]
\centering
\includegraphics[scale=0.22]{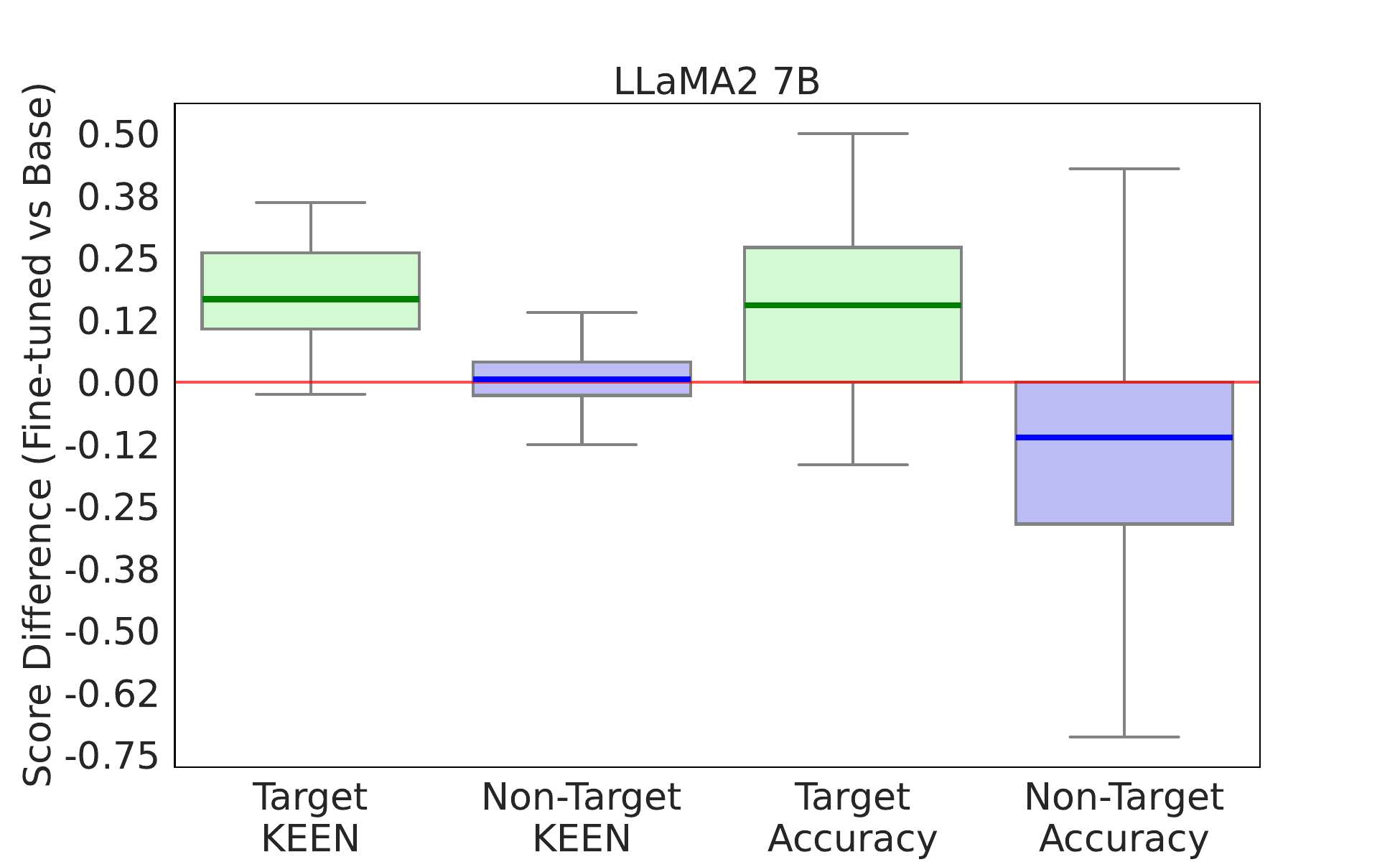}
\caption{Changes in the \method{} QA score and average QA accuracy after fine-tuning LLaMA2 7B on paragraphs about a target subject. These results are aggregated over individual fine-tuning processes for 20 target subjects.}
\label{fig:litmus_llama2}
\end{figure}

Our experiments so far evaluated \method{} while keeping the underlying LLM fixed. A natural question that arises is whether changes in the model's knowledge are reflected in changes in the \method{} score. 
We test this by fine-tuning LLaMA2 7B on paragraphs about a target subject and measuring changes in both the average QA accuracy and the \method{} QA score.
Concretely, we sample 20 subjects from the QA test dataset and retrieve paragraphs from the Wikipedia page of each subject  using BM25 \cite{robertson1995okapi}.\footnote{We use the Wikipedia dump from August 28, 2023.} Then, we use LoRA \cite{hu2022lora} to fine-tune $< 0.5\%$ of the model's parameters, separately for each subject. 
After fine-tuning for a certain target subject, we compute the \method{} score for that subject, as well as for 256 non-target subjects from the QA test dataset. The \method{} QA probe trained over the model's hidden states before fine-tuning is used to compute these scores.

Figure~\ref{fig:litmus_llama2} shows that on average, QA accuracy scores for the target entities increase by 0.16 and \method{} QA scores increase by 0.18 as models are fine-tuned on paragraphs related to them. Conversely, inline with works about catastrophic forgetting which find that models tend to forget information about entities observed in pre-training \citep{tirumala2022memorization}, the QA accuracy scores for non-target entities decrease after fine-tuning. However, the \method{} scores for non-target entities stay relatively constant. Since fine-tuning often doesn't erase residual information in LLMs \citep{patil2024can, hong2024intrinsic}, and \method{} relies on intermediate representations, a possible explanation for this discrepancy is that information is still encoded in the representations but the model fails to recall it.
To test this hypothesis, we selected 85 non-target entities whose decrease in QA accuracy after fine-tuning, $0.50 \pm 0.18$ to $0.24 \pm 0.15$, was not reflected in changes in their \method{} scores post fine-tuning, $0.46 \pm 0.12$ to $0.49 \pm 0.12$. We then use activation patching to recover QA accuracy from their intermediate fine-tuned representations, demonstrating that they still encode entity-related information \citep[]{ghandeharioun2024patchscopes, mosbach-etal-2020-interplay}.  

\paragraph{Skip fine-tuned layers to recover accuracy (FT subj.)} This experiment shows that information is recoverable from the fine-tuned representations when later layers in the fine-tuned model are bypassed. The procedure involves feeding the fine-tuned model a question, extracting only the \textit{subject representations} (for all subject tokens) from an intermediate later, patching them into the penultimate layer of the \textit{fine-tuned} model, and then measuring the patched QA accuracy per subject. Patching increases accuracy by $0.24 \pm 0.18$ above fine-tuned accuracy (FT) and restores accuracy within $-0.01 \pm 0.17$ of pre-trained accuracy (PT), per subject on average.

\paragraph{Apply pre-trained layers to recover accuracy (PT layer)} This experiment shows that information can still be extracted from fine-tuned representations using the pre-trained model. The procedure involves feeding the fine-tuned model a question, extracting \textit{all representations} from an intermediate layer, patching them into the penultimate layer of the \textit{pre-trained} model, and then measuring the patched QA accuracy per subject. Similarly, patching increases accuracy by $0.27 \pm 0.19$ above FT and restores accuracy within $-0.03 \pm 0.19$ of PT. 

These experiments demonstrate that \method{} scores are reflective of the knowledge encoded in the intermediate representations, and that the estimation gap between \method{} scores and QA accuracy post fine-tuning is because the later layers in the fine-tuned model suppress the generation of this information. Further details about the patching procedure and results are presented in \S\ref{subsec:recover_qa}.


\begin{figure}[t]
  \centering
   \includegraphics[scale=.26]{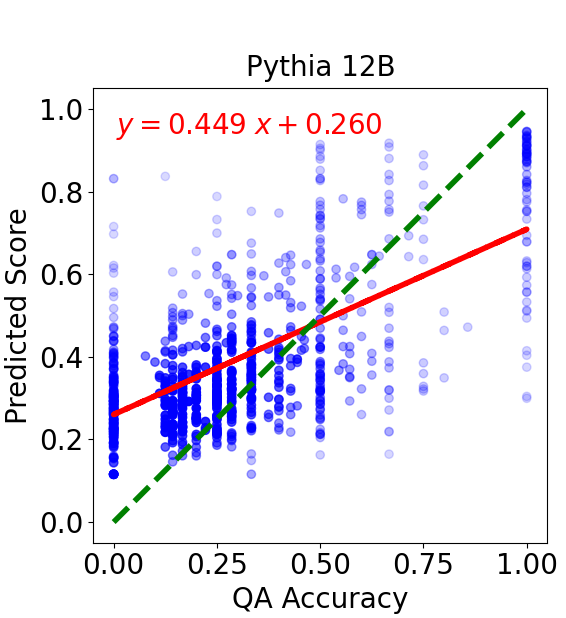}
   \includegraphics[scale=.26]{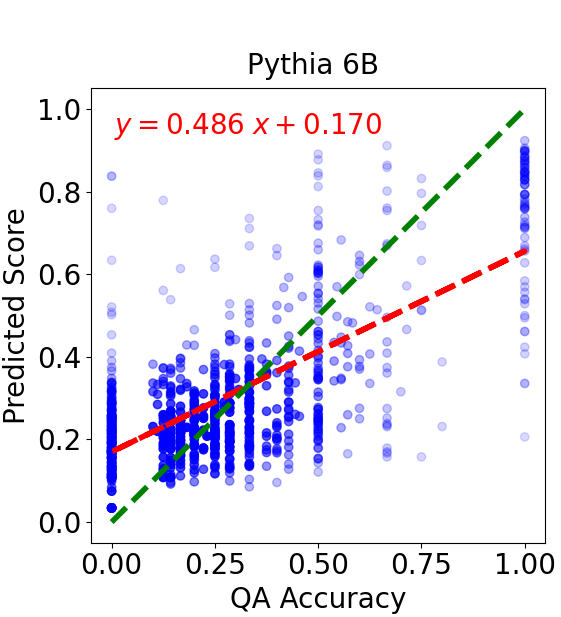}
   \includegraphics[scale=.26]{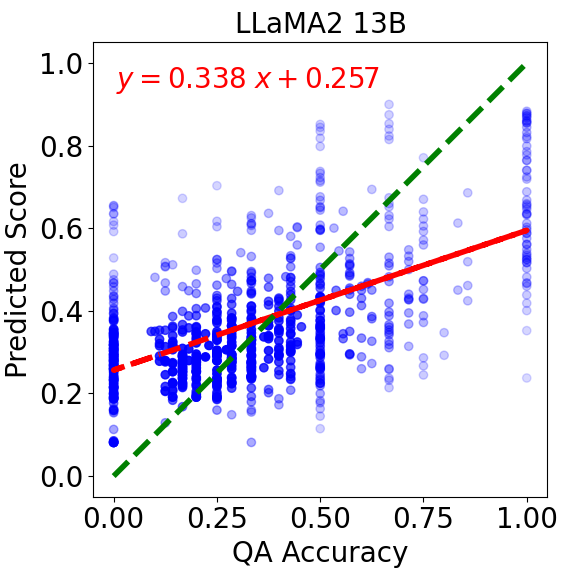}
   \includegraphics[scale=.26]{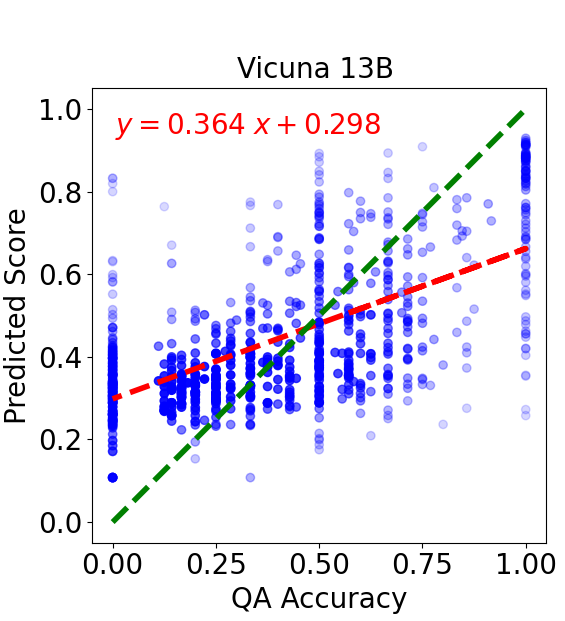}
  \caption{Predicted scores from the \method{} QA VP probe and the golden QA Accuracy scores are positively linearly related.}
  \label{fig:qa_probe_scatterplots}
\end{figure}

\begin{figure}[t]
  \centering
   \includegraphics[scale=.26]{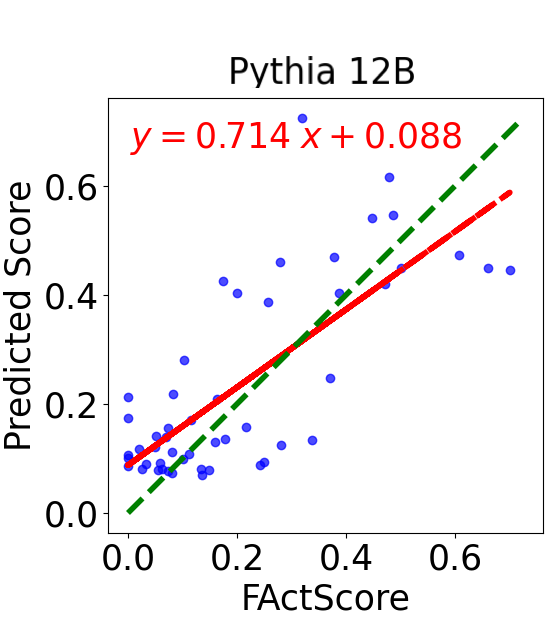}
   \includegraphics[scale=.26]{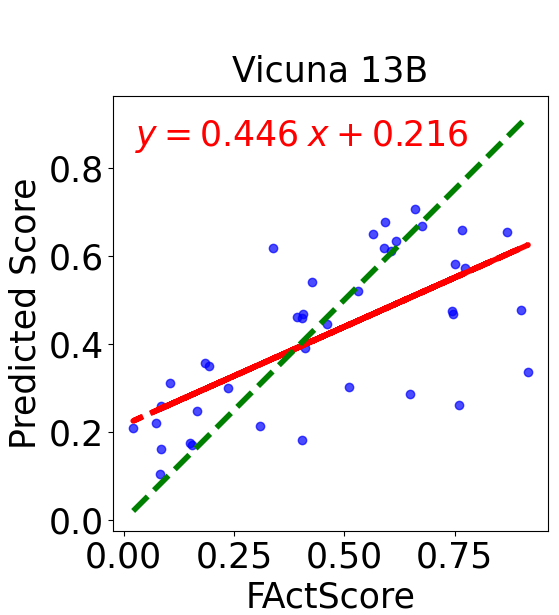}
  \caption{Predicted scores of the \method{} OEG VP probe versus FActScore scores. \method{} scores are positively linearly correlated with FActScore scores.}
  \label{fig:factscore_probes_corr_factscore_both}
\end{figure}

\subsection{Error Analysis}
\label{subsec:error_analysis}

To better understand the limitations of \method{}, we plot the probes' predicted scores against the reference QA accuracy and FActScore scores. 

Figure \ref{fig:qa_probe_scatterplots} shows that the \method{} VP QA probes tend to predict higher scores relative to QA accuracy for subjects that the model knows less about, although the \method{} scores for entities with QA accuracy between $[0, 0.5]$ do generally fall within a similar range of $[0.1, 0.5]$. For subjects that the model is more knowledgeable about, \method{} QA scores are more conservative, as seen by the cluster of scores below the $y=x$ line for QA accuracy values close to 1.0. Generally, \method{} scores have less variance than the QA accuracy scores since the slopes of the trend-lines are $<1$, which may suggest that more complex predictors are needed to capture all the variance of QA accuracy. These trends are consistent across models of different families and sizes. 

In \S\ref{subsec:qa_corr_plots_more} and \S\ref{subsec:oeg_corr_plots_more}, we include results for the other models, which follow the same trends in Figure~\ref{fig:qa_probe_scatterplots} and Figure~\ref{fig:factscore_probes_corr_factscore_both}. We also provide the scatter plots for probes trained on the different \method{} features and baselines, all demonstrating the same linear relations.

\subsection{Feature Analysis for VP-25 and VP-50}
\label{subsec:feature_analysis}
\begin{figure}[t]
  \centering
   \includegraphics[scale=.16]{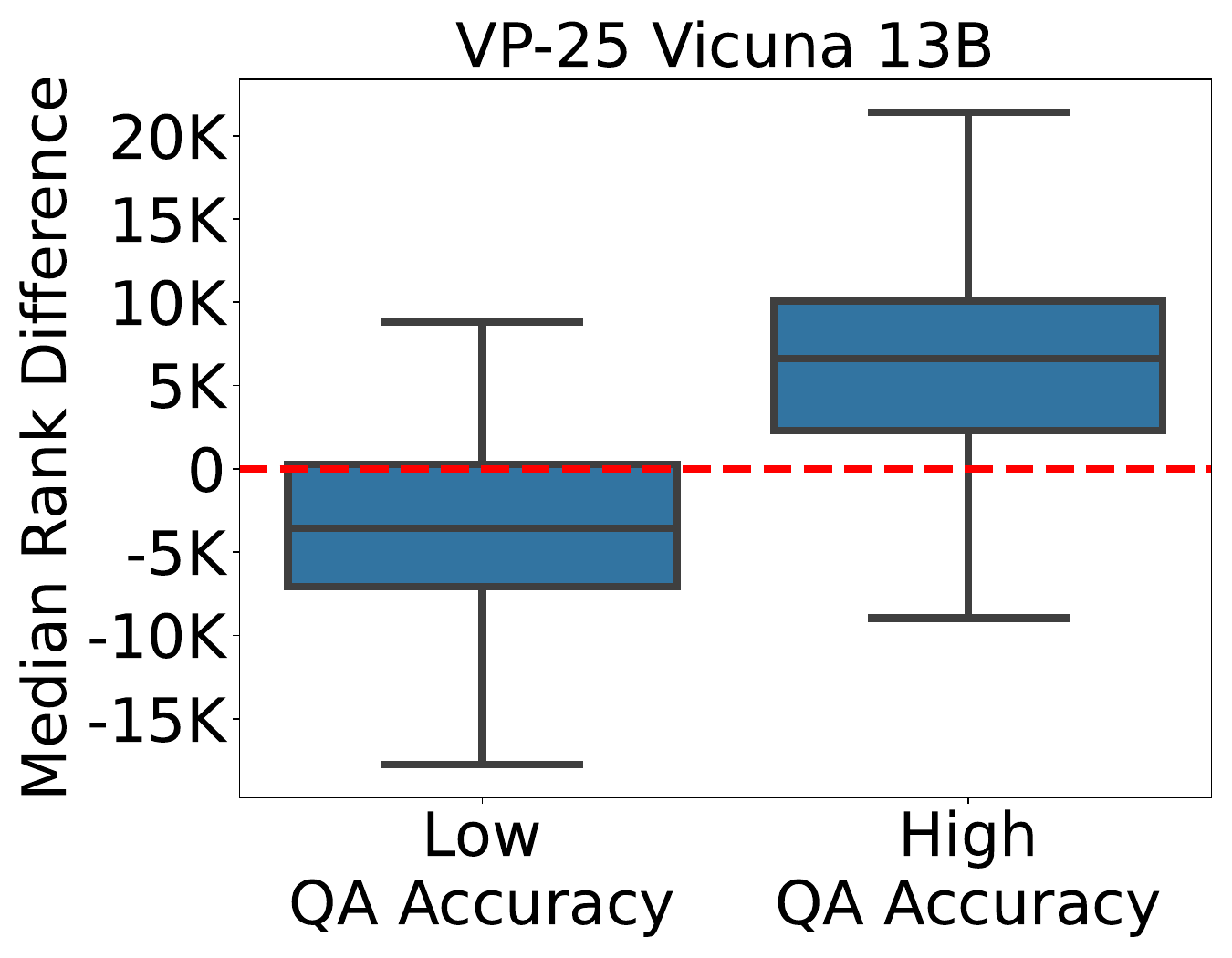}
   \includegraphics[scale=.16]{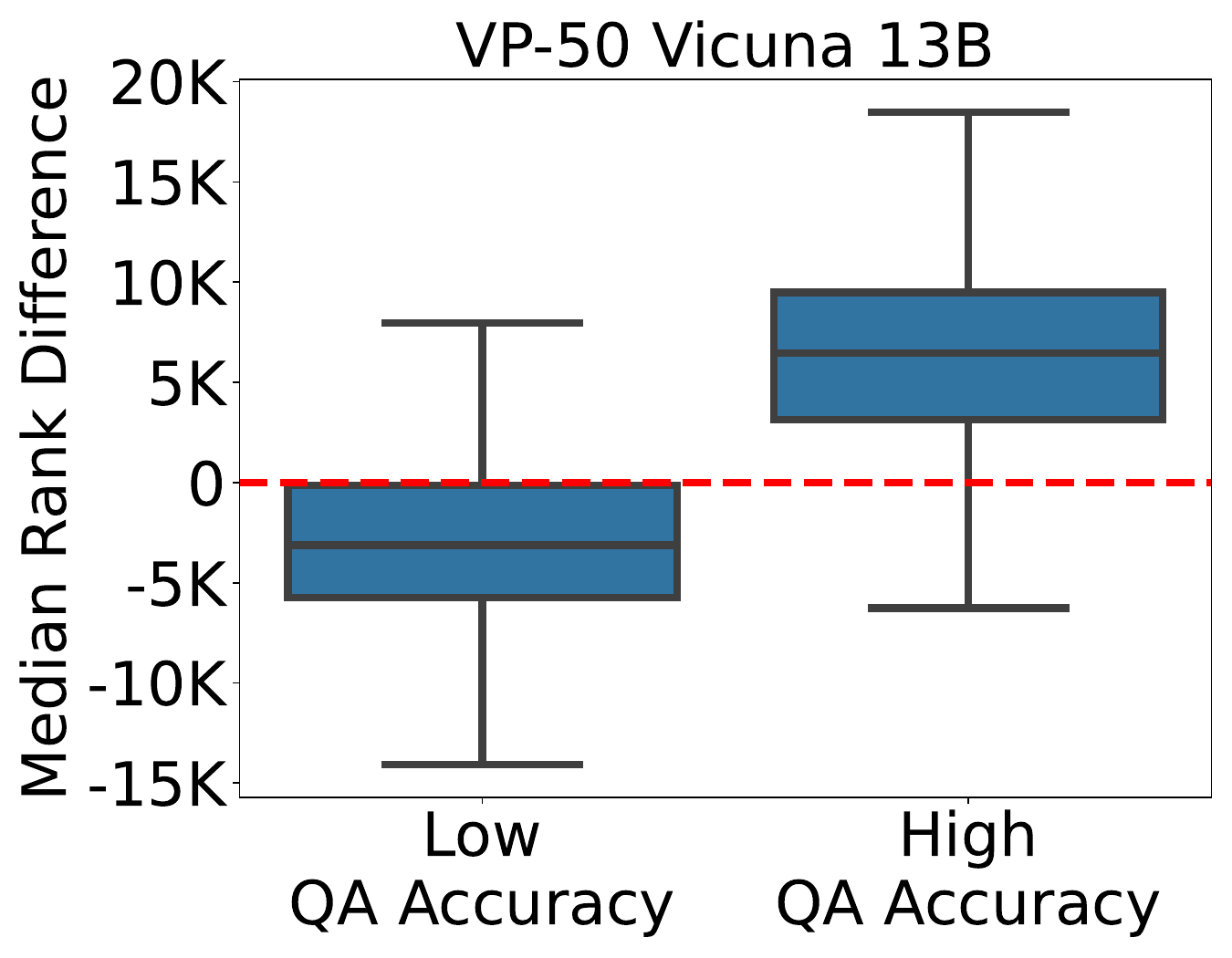}
   \includegraphics[scale=.16]{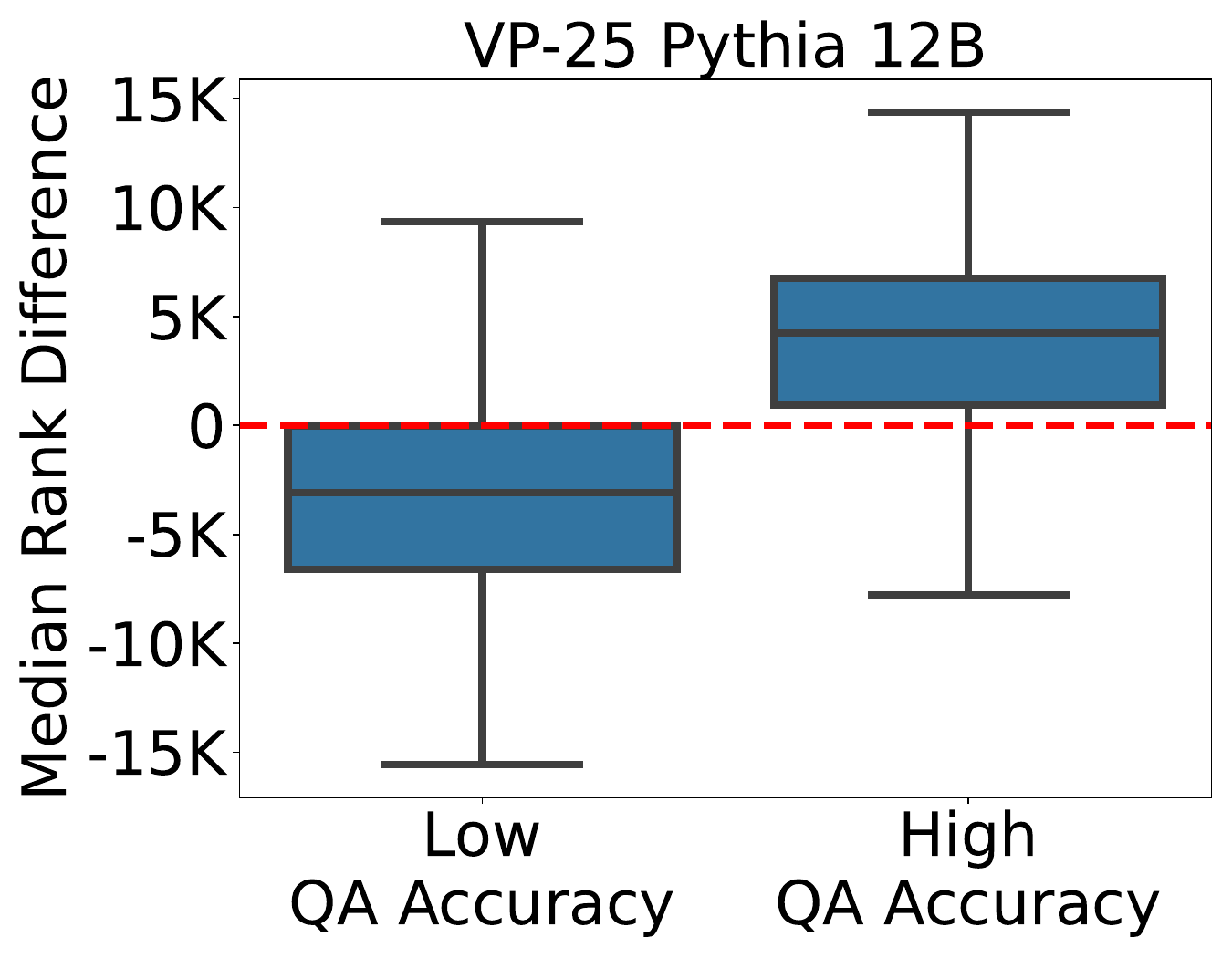}
   \includegraphics[scale=.16]{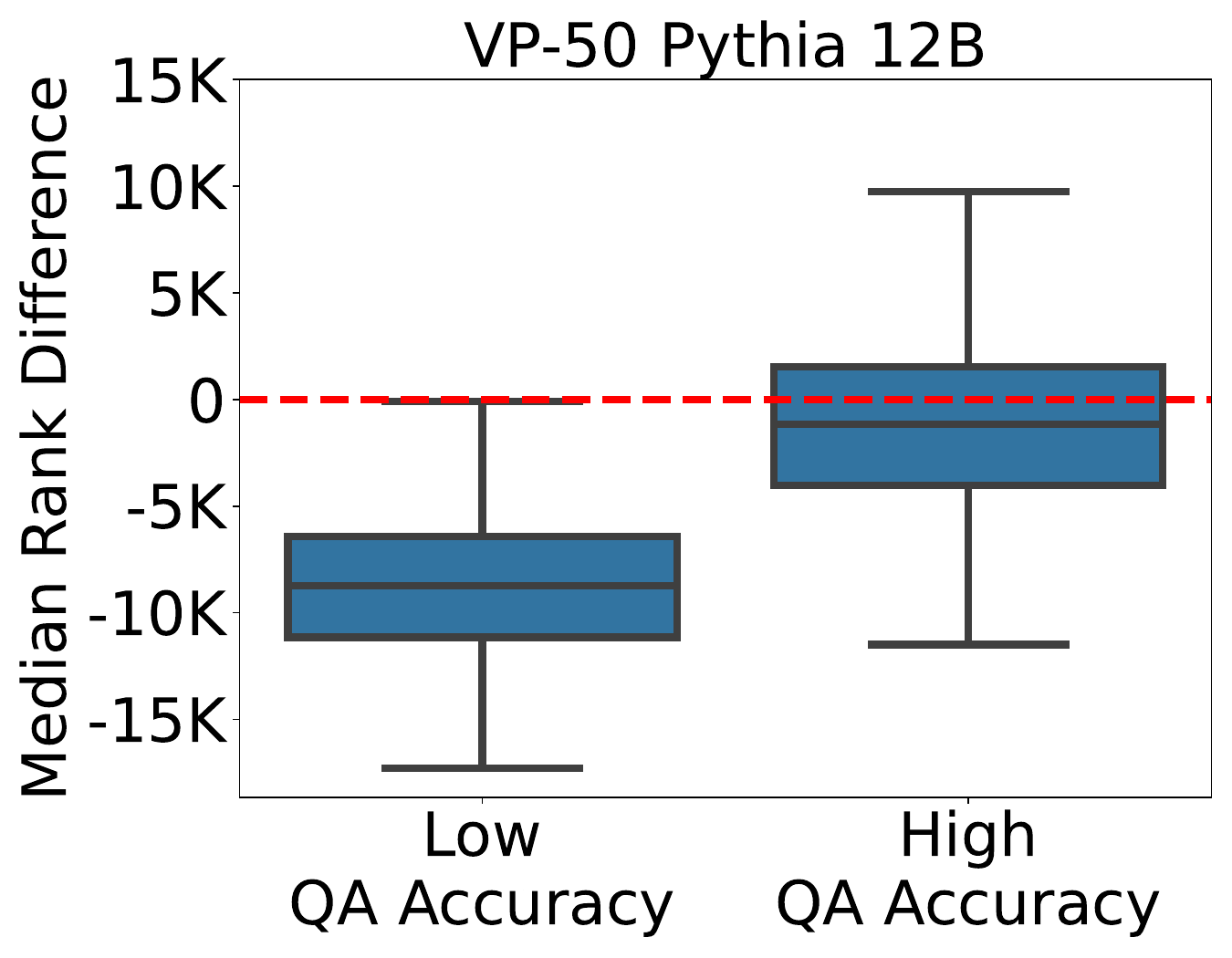}
  \caption{Difference, per subject, in the median rank of tokens with negative weight and tokens with positive weight.  Pythia 12B VP-25 and VP-50 show the trade-off between interpretability and performance -- though the median ranks of negative weight tokens are higher on average than positive weight tokens in VP-50, there is still a clear split in both accuracy groups.}
  \label{fig:vocab_proj_analysis}
\end{figure}

\begin{table}[t]
\footnotesize
\setlength\tabcolsep{3.8pt}
    \centering
    \resizebox{\linewidth}{!}{
    \begin{tabular}{lcp{6.5cm}}
    \toprule
          & Weight & Example influential tokens \\
         \midrule
         \multirow{4.5}{1em}{\rotatebox{90}{Pythia 12B}} & \multirow{2}{*}{Pos} & {\texttt{analysis, Statistical, Players, Senator, Quantum, nationality, investments}} \\ \cmidrule{2-3}
          & \multirow{2}{*}{Neg} & {\texttt{circadian, AMPK, lys, 16, jo, VERT, diese, see, Mort, ))*, dep, imi, ac }}\\
         \midrule 
         \multirow{4.5}{1em}{\rotatebox{90}{Vicuna 13B}} & \multirow{2}{*}{Pos} & {\texttt{athlet, kick, swing, developer, compiling, official, sales, GitHub, Movie}}{\texttt{}}\\ \cmidrule{2-3}
          & \multirow{2}{*}{Neg} & {\texttt{sle, hurt, Circ, Alt, book, JK, ja, adow, istema, ppings, adjust, istol}}\\
         \bottomrule
    \end{tabular}
    }
    \caption{Examples of the most influential tokens in the \method{} QA VP probes that were assigned positive and negative weights. These are some of the tokens that correspond to the features of \method{} QA VP-50 probes.}
    \label{tab:vocab_proj_examples}
\end{table}

\paragraph{Identifying prominent features}
We analyze the most influential features of the \method{} QA VP probes to understand which tokens contribute most to predicting average QA accuracy. Our goal in this analysis is to identify tokens that either increase or decrease predicted QA accuracy, and to determine whether they are promoted in the representations of subjects with high and low QA accuracy, respectively. As a concrete example, for subjects with low QA accuracy, we expect tokens that decrease QA accuracy to generally be ranked higher in the subject representations than tokens that increase QA accuracy (the highest rank is 0 which corresponds to the token with highest logit value). Since the input normalization scheme described in \S\ref{subsec:features} normalizes a token's logit in a given hidden state by its magnitude across all subjects (not with respect to the other tokens in the hidden state), we can interpret the weight learned by the \method{} QA VP probe for each token as its direction and magnitude of influence on the predicted score.

First, we identify the tokens associated with the largest absolute weights in the \method{} QA VP probes, as they are most influential on the predicted score. Next, we compare the median rank of tokens with negative weights to those with positive weights in the vocabulary projections of subjects with high QA accuracy (1.0) and low QA accuracy (0.0).
Figure~\ref{fig:vocab_proj_analysis} shows that for low QA accuracy subjects, the median rank of negative weight tokens is generally higher than that of positive weight tokens. Conversely, for high QA accuracy subjects, the median rank of negative weight tokens is generally lower than that of positive weight tokens. This opposing trend in the two accuracy groups indicates that there is a small set of tokens which hold signals for differentiating between subjects the model knows more about and those it knows less about. We provide these important tokens in Table~\ref{tab:vocab_proj_examples}. 

\paragraph{Mapping knowledge clusters and ``holes''}
Tokens assigned positive weight are related to meaningful concepts while tokens assigned negative weight are often numbers, abbreviations, or suffixes. A possible interpretation of this difference is that the hidden states of low accuracy subjects reflect ``holes'' in the model's knowledge, thereby encoding less information and promoting less semantically meaningful tokens. These negative weight tokens may help identify additional ``holes'' if similarly promoted in other subject representations. Conversely, positive weight tokens can be used to identify clusters of knowledge. Table~\ref{tab:pos_tok_analysis} shows that Pythia 12B and Vicuna 13B encode more information about political figures, athletic players, or movies. This is evidenced by the $0.20-0.55$ higher than average QA accuracy for a sample of subjects with high logit values ($\geq 0.65$) for tokens like \texttt{Senator}, \texttt{Player}, \texttt{Movie}, and \texttt{athlet}.

\begin{table}[t]
\setlength{\tabcolsep}{3.8pt}
\footnotesize
\centering
\resizebox{\linewidth}{!}{
\begin{tabular}{lclcc}
    \toprule
    & {Token} & {Subject} & {Logit} & {QA}\\
    \midrule
    \multirow{10.5}{*}{\rotatebox[origin=c]{90}{Pythia 12B}} & \multirow{5}{*}{\texttt{Senator}} & {Rand Paul} & 0.84 & 0.55\\
    & & {Amin Amidu Sulemana} & 0.79 & 0.67\\
    & & {Beau Biden} & 0.67 & 0.54\\
    & & {Chelsea Clinton} & 0.68 & 0.58\\
    & & \textit{Mean Entity} & \textit{0.41} & \textit{0.35}\\
    \cmidrule{2-5}
    & \multirow{5}{*}{\texttt{Players}} & {Charlton Athletic F.C.} & 0.84 & 0.60\\
    & & {Aston Villa F.C.} & 0.81 & 0.60\\
    & & {AC Milan} & 0.79 & 0.75\\
    & & {Moneyball} & 0.76 & 0.67\\
    & & \textit{Mean Entity} & \textit{0.44} & \textit{0.35}\\
    \midrule
    \multirow{10.5}{*}{\rotatebox[origin=c]{90}{Vicuna 13B}} & \multirow{5}{*}{\texttt{Movie}} & {Attarintiki Daredi} & 0.86 & 0.71 \\
    & & {Iron Man 3} & 0.83 & 0.63\\
    & & {Friday the 13th Part 2 } & 0.76 & 0.88\\
    & & {National Treasure: Book of Secrets} & 0.76 & 0.86\\
    & & \textit{Mean Entity} & \textit{0.31} & \textit{0.45}\\
    \cmidrule{2-5}
    & \multirow{5}{*}{\texttt{athlet}} & {Charlotte Hornets} & 0.97 & 1.00 \\
    & & {Chariots of Fire} & 0.84 & 0.63 \\
    & & {Olympique Lyonnais} & 0.74 & 0.60 \\
    & & {The Cowboys} & 0.67 & 0.71 \\
    & & \textit{Mean Entity} & \textit{0.41} & \textit{0.45}\\
    \bottomrule
\end{tabular}}
\caption{Examples of subjects from the full validation set with high logit values for influential positive tokens. Their high QA accuracies, relative to the \textit{Mean Entity}, indicate knowledge clustered around the specific concepts: political figures, athletes/players, and movies.}
\label{tab:pos_tok_analysis}
\end{table}

\section{Related Work}   

\paragraph{Evaluation of knowledge and factuality of LLMs}
The common practice for estimating knowledge in LLMs is to query the model and then evaluate its outputs. This is often conducted though question-answering setups with gold labels \citep[][inter alia]{roberts-etal-2020-much, Petroni2019LanguageMA, cohen-etal-2023-crawling}, by letting the model generate multiple responses and measuring response consistency \citep{cohen-etal-2023-lm, manakul-etal-2023-selfcheckgpt, kuhn2023semantic}, checking whether the generated output is supported by external evidence \cite{gao-etal-2023-rarr, bohnet2022attributed, min-etal-2023-factscore}, or by estimating the model's uncertainty per-response \cite{zhang-etal-2023-enhancing-uncertainty, jesson2024estimating}. 
Unlike these methods, we focus on evaluating the model's entity knowledge beyond a single response, based on intrinsic features extracted before generating a single token.

\paragraph{Probing internal representations of LLMs}
Probing over internal representations has been used to predict model behavior, such as truthfulness \citep{marks2024the, azaria2023the}, and properties of language, such as part-of-speech \citep{belinkov-etal-2017-neural, nikolaev-pado-2023-investigating}, syntax \cite{hewitt-manning-2019-structural}, and sentence length \citep{adi2017finegrained} \textit{for a specific input}. Probing has also been used to identify which hidden states are most influential on the performance of tasks, like classification \citep{alain2017understanding}. Our use of probing differs from prior work because we estimate a property that captures model behavior over many inputs rather than a single input. Namely, the \method{} score provides a knowledge estimate relevant to any input concerning the entity. Further, \method{} focuses on estimating entity-specific knowledge and is useful in evaluating several model behaviors, including hedging, shifts in knowledge, and truthfulness.

\paragraph{Hallucination detection using intrinsic features}
Our work is closely related to methods that leverage intrinsic features for detecting factually incorrect claims, but has two core differences. The first being in our choice of features: we specifically use the hidden states corresponding to the named entity from the upper intermediate layers. In contrast, existing methods use various other features, like intermediate activation values \citep{azaria-mitchell-2023-internal}, outputs from the self-attention modules \cite{yu2024mechanisms, yuksekgonul2024attention, li2023inferencetime, snyder2023early}, soft-max prediction probabilities, and fully-connected scores \citep{snyder2023early}. \citet{yu2024mechanisms, goloviznina2024ive, su2024unsupervised} also examine the intermediate hidden representations, but for the purpose of identifying whether there exists a subspace of hidden states that lead to hallucinations. Similarly to our work, \citet{yu2024mechanisms} uses the hidden representations of subjects, but rather to train a binary hallucination detector. Unlike all these works that use internal representations to predict the factuality of a specific claim, we learn to estimate knowledge from a single internal representation of an entity, which is applicable to any claim pertaining to it.

\section{Conclusion}    
We present the problem of estimating entity knowledge solely from the model's internal representations of the entity. We show that \method{} offers a simple and interpretable solution which correlates with model performance in both QA and OEG settings, as well as with current hallucination detection methods. Further, \method{} is also reflective of both hedging behavior and changes in knowledge throughout fine-tuning.
From a broad perspective, our results demonstrate the potential of estimating model qualities and behavior for certain inputs based on intrinsic features, and call for future work to leverage simple and efficient methods like \method{} to improve the factuality and reliability of LLMs.

\section*{Limitations}
While our approach successfully estimates the extent of the model's knowledge about a subject, it does not identify the presence or lack of knowledge about specific facts. For instance, \method{} can estimate that the model will be 55\% truthful when generating content about \textit{Napoleon}, but it does not pinpoint that the model is unable to answer the specific question, \textit{What military academy did Napoleon attend?}. An interesting direction for future work would be to develop a more fine-grained approach that predicts how knowledgeable the model is about specific aspects of the subject (e.g. military career of Napoleon) or identifies specific facts encoded in subject representations.

Another limitation is that this work focuses on estimating knowledge for entities, however not all subjects of questions are entities. For example, there is no clear subject for which we can apply \method{} in the question, \textit{How does exercise influence mental health?}. \method{} also assumes that the subjects are already extracted for analysis. While identifying named entities in text is a well-studied task in NLP \citep{Nadeau2007ASO}, 
combining it with \method{} could make this approach more complex and computationally expensive.

Our evaluation focuses only on transformer-based auto-regressive LLMs. While this is one of the most popular and largest families of LLMs, it would be valuable to study the applicability of \method{} to other model architectures. Notably, \citet{sharma2024locating} shows that factual recall in Mamba is similarly centered in the hidden states of the last subject token from the intermediate layers, so we expect our approach to generalize to other recurrent architectures.

\section*{Acknowledgements}
We thank Amir Globerson for insightful discussions and feedback throughout the project. We also thank Ori Yoran and Gal Yona for providing useful feedback on this manuscript. This work was supported by the Tel Aviv University Center for AI and Data Science (TAD), the Israeli Science Foundation, and the Deutsch Foundation.

\bibliography{custom,anthology}

\clearpage
\appendix

\section{\method{} Training Details}
\label{sec:training_details}

\subsection{Hyper-parameter Tuning and Resources}
\paragraph{Hyper-parameter tuning for \method{} probes}
All \method{} QA and OEG probes were trained with the AdamW optimizer with weight decay 0.01, and batch size of 32, and learning rates were optimized over $1e^{-3}, 5e^{-3}, 5e^{-4}, 1e^{-4}, 1e^{-5}, 5e^{-5}$. The best hyper-parameters for QA and OEG \method{} probes are found in Table~\ref{tab:qa_training_hyperparameters} and Table~\ref{tab:oeg_training_hyperparameters}, respectively. The epochs represents the maximum number of epochs set for training since we did not implement early stop, however all evaluations are done on the checkpoint with the highest Pearson correlation on the validation set.

\begin{table}[ht] 
\centering
\scriptsize
\setlength\tabcolsep{1.8pt}
\begin{tabular}{SSSSSSSS}
\toprule
    {\method{}} & {Hyper} & {GPT2} & {Pythia} & {Pythia} & {LLaMA} & {LLaMA} & {Vicuna}\\ 
    {Probe} & {Param} & {XL} & {6B} & {12B} & {7B} & {13B} & {13B}\\ 
    \midrule
    {HS} & {Epoch} & {3K} & {100} & {100} & {1K} & {1K} & {100} \\
    {} & {LR} & {$1e^{-5}$} & {$1e^{-4}$} & {$1e^{-4}$} & {$1e^{-5}$} & {$1e^{-5}$} & {$1e^{-4}$} \\
    \midrule
    {VP} & {Epoch} & {3K} & {100} & {3K} & {1K} & {1K} & {3K} \\
    {} & {LR} & {$1e^{-5}$} & {$1e^{-4}$} & {$1e^{-5}$} & {$1e^{-5}$} & {$1e^{-5}$} & {$1e^{-5}$} \\
    \midrule
    {VP-200} & {Epoch} & {3K} & {500} & {3K} & {3K} & {3K} & {3K} \\
    {} & {LR} & {$1e^{-5}$} & {$1e^{-4}$} & {$1e^{-4}$} & {$1e^{-4}$} & {$1e^{-4}$} & {$1e^{-4}$} \\
    \midrule
    {VP-100} & {Epoch} & {3K} & {500} & {3K} & {3K} & {3K} & {3K} \\
    {} & {LR} & {$1e^{-5}$} & {$1e^{-4}$} & {$1e^{-4}$} & {$1e^{-4}$} & {$1e^{-4}$} & {$1e^{-4}$} \\
    \midrule
    {VP-50} & {Epoch} & {3K} & {1K} & {3K} & {3K} & {3K} & {3K} \\
    {} & {LR} & {$1e^{-5}$} & {$1e^{-4}$} & {$1e^{-4}$} & {$1e^{-4}$} & {$1e^{-4}$} & {$1e^{-4}$} \\
    \midrule
    {VP-25} & {Epoch} & {5K} & {3K} & {3K} & {3K} & {3K} & {3K} \\
    {} & {LR} & {$1e^{-5}$} & {$1e^{-4}$} & {$1e^{-4}$} & {$1e^{-4}$} & {$1e^{-4}$} & {$1e^{-4}$} \\
    \midrule
    {VP-10} & {Epoch} & {-} & {-} & {-} & {3K} & {-} & {-} \\
    {} & {LR} & {$1e^{-5}$} & {$1e^{-4}$} & {$1e^{-4}$} & {$1e^{-4}$} & {$1e^{-4}$} & {$1e^{-4}$} \\
    \midrule
    {FC} & {Epoch} & {-} & {-} & {-} & {1K} & {-} & {-} \\
    {} & {LR} & {-} & {-} & {-} & {$1e^{-5}$} & {-} & {-} \\
    \midrule
    {ATTN} & {Epoch} & {-} & {-} & {-} & {1K} & {-} & {-} \\
    {} & {LR} & {-} & {-} & {-} & {$1e^{-5}$} & {-} & {-} \\
    \bottomrule
\end{tabular}
\caption{Hyper-parameters for \method{} QA probes and baselines.}
\label{tab:qa_training_hyperparameters}
\end{table}

\begin{table}[t] 
\centering
\scriptsize
\renewcommand{\arraystretch}{.1}
\begin{tabular}{SSSS}
\toprule
    {\method{} Probe} & {Hyper-parameter} & {Pythia 12B} & {Vicuna 13B}\\ 
    \midrule
    {HS} & {Epoch} & {1K} & {1K} \\
    {} & {LR} & {$1e^{-5}$} & {$1e^{-4}$} \\
    \midrule
    {VP} & {Epoch} & {5K} & {3K} \\
    {} & {LR} & {$1e^{-5}$} & {$1e^{-5}$} \\
    \midrule
    {VP-200} & {Epoch} & {5K} & {3K} \\
    {} & {LR} & {$1e^{-4}$} & {$1e^{-4}$} \\
    \midrule
    {VP-100} & {Epoch} & {5K} & {5K} \\
    {} & {LR} & {$1e^{-4}$} & {$5e^{-4}$} \\
    \midrule
    {VP-50} & {Epoch} & {5K} & {5K} \\
    {} & {LR} & {$1e^{-4}$} & {$5e^{-4}$} \\
    \midrule
    {VP-25} & {Epoch} & {5K} & {5K} \\
    {} & {LR} & {$5e^{-4}$} & {$1e^{-3}$} \\
    \midrule
    {VP-10} & {Epoch} & {5K} & {5K} \\
    {} & {LR} & {$1e^{-3}$} & {$5e^{-4}$} \\
    \midrule
    {FC} & {Epoch} & {1K} & {1K} \\
    {} & {LR} & {$5e^{-5}$} & {$5e^{-5}$} \\
    \midrule
    {ATTN} & {Epoch} & {1K} & {1K} \\
    {} & {LR} & {$1e^{-5}$} & {$5e^{-5}$} \\
    \bottomrule
\end{tabular}
\caption{Hyper-parameters for \method{} OEG probes and baselines.}
\label{tab:oeg_training_hyperparameters}
\end{table}

\paragraph{Hyper-parameters for fine-tuning LLaMA2 7B}

\begin{table}[ht] 
\centering
\scriptsize
\setlength\tabcolsep{1.8pt}
\begin{tabular}{SSSSSSSS}
\toprule
    {Optimizer} & {LR} & {Epoch} & {Scheduler} & {Warm Up} & {LoRA} & {LoRA} &{LoRA}\\ 
    {} & {} & {} & {} & {ratio} & {alpha} & {dropout} & {r}\\ 
    \midrule
    {AdamW} & {$2e^{-4}$} & {100} & {Linear} & {0.03} & {16} & {0.1} & {64}\\ 
    \bottomrule
\end{tabular}
\caption{Hyper-parameters for fine-tuning LLaMA2 7B.}
\label{tab:lora_training_hyperparameters}
\end{table}

The training details for the fine-tuning experiment in \S\ref{subsec:training} are described in Table~\ref{tab:lora_training_hyperparameters}.

\paragraph{Resources}
All our experiments were conducted using the PyTorch package \citep{NEURIPS2019_bdbca288} on a single A100 or H100 GPU.

\subsection{Choice of Input Layer Configuration}
\label{subsec:layer_config}
Our choice of choosing representations from the upper middle layers and the last subject position were motivated by previous work \citep[]{geva-etal-2023-dissecting, meng-locating, hernandez2024inspecting}. The focus on the last subject position is because in auto-regressive models, earlier positions cannot capture the full subject name. To systematically defend this decision, we repeated experiments recorded in Table~\ref{tab:factscore_corr_qa} with LLaMA2 7B and Pythia 12B and trained new \method{} QA VP probes using the following layers as features: first 3 layers (Early) , last 3 layers (Late), 1 layer upper-intermediate layer (One) (22, 25), 5 upper-intermediate layers (Five) ([19, 24), [22, 27)). Correlation values between \method{} estimates and QA accuracy are statistically significant ($p \leq 2e^{-50}$), and provided in Table~\ref{tab:layer_choice}.

\begin{table*}[t]
\centering
\begin{tabular}{lccccc}
\toprule
Model        & Early & Late & One  & Five & \method{} VP \\
\midrule
Pythia 12B   & 0.50  & 0.64 & 0.60  & 0.64 & \textbf{0.64} \\
LLaMA2 7B    & 0.45  & 0.63 & 0.64 & 0.64 & \textbf{0.64} \\
\bottomrule
\end{tabular}
\caption{Pearson correlation values of QA probes trained on hidden representations from various layer configurations.}
\label{tab:layer_choice}
\end{table*}

We find a clear benefit to using upper-intermediate to late layers over early layers and no improvement when averaging over multiple upper intermediate layers. Using the late layers has comparable correlation values as using the upper intermediate layers, which supports previous findings that knowledge is aggregated in the subject's hidden states of the upper intermediate layers \citep[]{geva-etal-2023-dissecting, meng-locating}.

\subsection{Discussion of \method{} Efficiency and Training Time Statistics}
\method{} has two key efficiency advantages. (1)~Though \method{} probes are only trained on a small set of entities, they generalize well, which is especially significant considering that models are trained on millions of entities. Estimating knowledge with \method{}, for subjects not used to train the probes, requires a single inference pass and the execution of the \method{} probe. In contrast, the computational cost of query-based evaluation requires multiple inference passes per query. With every query requiring 1-5 passes (due to tokenization) and an average number of 5.3 queries per subject, query-based evaluation takes 5.3-27 times longer. (2) \method{} can be applied ubiquitously to all subjects since it does not require labeled data, beyond a small sample for training. Labeling is a major hurdle, especially for OEG where complex processing and verification of the output is needed \citep{min-etal-2023-factscore}. Query-based evaluation approaches also require access to knowledge bases (KB) which inherently restricts the subjects that can be evaluated to those in the KB.

To demonstrate the training time benefits of \method{}, in Table~\ref{practical_train_qa} we report the epoch and practical length of training for \method{} QA probes used in our evaluations.

\begin{table}[ht]
\footnotesize{}
\centering
\caption{Practical number of training epochs and times of \method{} QA probes.}
\label{practical_train_qa}
\begin{tabular}{lccc}
\toprule
\multirow{2}{*}{Model}   & \multirow{2}{*}{Probe} & {Best} & {Training} \\
   & {} & {Epoch} & {(Min)} \\ \midrule
\multirow{2}{*}{GPT2 XL}   & VP  & 110   & 5    \\
                           & HS  & 3K    & 20   \\ \midrule
\multirow{2}{*}{Pythia 12B} & VP  & 279   & 30   \\
                           & HS  & 100   & 1    \\ \midrule
\multirow{2}{*}{Pythia 6B}  & VP  & 44    & 2.5  \\
                           & HS  & 100   & 8    \\ \midrule
\multirow{2}{*}{LLaMA2 7B}  & VP  & 167   & 5    \\
                           & HS  & 1K    & 10   \\ \midrule
\multirow{2}{*}{LLaMA2 13B} & VP  & 1K    & 1    \\ 
                           & HS  & 321   & 16.5 \\ \midrule
\multirow{2}{*}{Vicuna 13B} & VP  & 100   & 1    \\
                           & HS  & 202   & 5    \\
\bottomrule
\end{tabular}
\end{table}

\section{Dataset Details}
\label{sec:dataset_details}

\subsection{Question Answering Dataset}
Examples are sampled as (subject, predicate, object) triplets from Wikidata, and questions are formed using manually written templates with placeholders for the subject and answer entity type (from the “instance of” property in Wikidata). We form directed questions using answer entity types to decrease the ambiguity of the expected answer i.e. instead of ``Where was Barack Obama born?'' we ask ``In what city was Barack Obama born?''. Also, to increase the coverage of knowledge per entity, we extended the PopQA dataset to include an additional 26 relations to a total of 42 relations. Examples of answer templates and questions are found in Table~\ref{tab:relations_with_templates}. The train split consists of 2,223 entities and 12,324 questions, validation split has 555 entities and 3,003 questions, and test split has 694 entities and 3,821 questions.

\begin{table*}[!ht]
\centering
\resizebox{\textwidth}{!}{
\begin{tabular}{lll}
\midrule
{Relation} & {Question Template} & {Entity Count} \\
\midrule
genre                          & What genre is [subj]? & 1979 \\ 
country of origin              & What is the country of origin of [subj]? & 1574 \\ 
director                       & Who was the director of [subj]? & 1196 \\
screenwriter                   & Who was the screenwriter of [subj]? & 1174 \\
producer                       & Who was the producer of [subj]? & 1163 \\ 
occupation                     & What is [subj]'s occupation? & 1092 \\
color                          & What color is [subj]? & 1044 \\
composer                       & Who was the composer of [subj]? & 1041 \\
place of birth                 & In what [obj\_type] was [subj] born? & 977 \\ 
country of citizenship         & What is [subj]'s country of citizenship? & 922 \\
country                        & In what country is [subj]? & 913 \\
languages spoken, written or signed & What language does [subj] speak? & 632 \\ 
sport                          & What sport does [subj] play? & 503 \\
language of work or name       & What is the language of [subj]? & 493 \\
capital                        & What is the capital of [subj]? & 482 \\ 
author                         & Who is the author of [subj]? & 452 \\ 
performer                      & Who is the performer of [subj]? & 361 \\
educated at                    & What is the alma mater of [subj]? & 359 \\ 
place of death                 & In what [obj\_type] was [subj] born? & 343 \\ 
followed by                    & What [obj\_type] follows [subj]? & 332 \\ 
father                         & Who is the father of [subj]? & 327 \\ 
religion or worldview          & What is the religion of [subj]? & 276 \\ 
member of sports team          & What sports team does [subj] play for? & 270 \\
record label                   & What is the record label of [subj]? & 267 \\ 
mother                         & Who is the mother of [subj]? & 250 \\
position played on team / speciality & What sports position does [subj] play? & 241 \\ 
spouse                         & Who is the spouse of [subj]? & 218 \\ 
participant in                 & In what sports event did [subj] participate in? & 218 \\ 
publisher                      & Who is the publisher of [subj]? & 217 \\ 
sibling                        & Who is the sibling of [subj]? & 212 \\ 
child                          & Who is the child of [subj]? & 211 \\ 
capital of                     & What is [subj] the capital of? & 211 \\ 
native language                & What is the native language of [subj]? & 172 \\ 
religion or worldview          & What is the religion of [subj]? & 168 \\
member of political party      & What is the political party associated with [subj]? & 135 \\
work location                  & In what [obj\_type] does [subj] work in? & 110 \\ 
country for sport              & What country does [subj] play for? & 92 \\ 
headquarters location          & In what [obj\_type] are the headquarters of [subj] located? & 80 \\
league                         & What sports league does [subj] play in? & 74 \\ 
lyrics by                      & Who wrote the lyrics of [subj]? & 70 \\
consecrator                    & Who is the consecrator of [subj]? & 33 \\ 
editor                         & Who is the editor of [subj]? & 12 \\
\bottomrule
\end{tabular}
}
\caption{Templates used for generating questions for the QA dataset with the counts of attributed entities.}
\label{tab:relations_with_templates}
\end{table*}

\subsection{Open-Ended Generation Dataset}
The unlabeled 500 subject FActScore dataset was used to evaluate the \method{} OEG probes, and the transfer of \method{} QA probes to the OEG setting.

\subsection{Model Hedging Behavior}
\label{subsec:model_hedging_details}
The experiment described in \S\ref{subsec:hedging} assessed the correlation between the fraction of queries for which the model exhibited hedging behavior and the \method{} score for a given entity. To determine the hedging fraction, the model was prompted with a set of common question-answer pairs about the entity, and the proportion of responses containing an exact match with some hedging phrase was calculated. Hedging phrases were identified through manual analysis of model responses and included the expressions \texttt{``nobody knows,'' ``I'm sorry,'' ``I can't seem to find the answer,'' ``could you help me,'' ``can anyone help me,'' ``I'm not sure,'' ``I don't know,'' ``I'm not entirely sure,'' ``could you please provide more,'' ``could you provide more information,'' ``provide more context,''} and \texttt{``clarify your question.''}

\section{Additional Results}
\label{sec:additional_results}

\subsection{Mean Standard Error (MSE) for \method{}}
Table~\ref{tab:factscore_factscore_mse}, Table~\ref{tab:factscore_corr_mse}, Table~\ref{tab:qa_accuracy_mse}, present the MSE for the \method{} OEG probes with FActScore scores, \method{} OEG probes with FActScore scores, the \method{} QA probes with FActScore scores, and the \method{} QA probes with average QA accuracy scores. The performance of these probes is discussed in \S\ref{subsec:results}.
\begin{table}[ht] 
\centering
\footnotesize
\setlength\tabcolsep{2pt}
\begin{tabular}{SSSSSSS}
\toprule
    {Model} & {Freq.} & {FC} & {ATTN} & {VP-50} & {VP} & {HS} \\ \midrule
    {Pythia-12B} & {0.028} & {0.026} & {0.014} & {0.020} & {0.017} & {0.014} \\
    {Vicuna-13B} & {0.052} & {0.075} & {0.049} & {0.052} & {0.040} & {0.039} \\
    \bottomrule
\end{tabular}
\caption{MSE for \method{} OEG Probes between predicted \method{} scores and FActScore scores.}
\label{tab:factscore_factscore_mse}
\end{table}

\begin{table}[ht]
\centering
\footnotesize
\setlength\tabcolsep{3.5pt}
\begin{tabular}{lcccccc}
\toprule
    Model & Freq. & {FC} & {ATTN} & {VP-50} & {VP} & {HS} \\
    \midrule 
    Pythia 12B  & {-} & {0.053} & {0.043} & {0.047} & {0.062} & {0.074} \\ 
    Vicuna 13B  & {-} & {0.046} & {0.053} & {0.052} & {0.049} & {0.050} \\ \bottomrule
\end{tabular}
\caption{MSE for \method{} QA probes and FActScore. The MSE scores for FC are omitted because we do not compute \method{} probes on this feature.}
\label{tab:factscore_corr_mse}
\end{table}

\begin{table}[ht]
\centering
\footnotesize
\setlength{\tabcolsep}{2.7pt}
\begin{tabular}{lSSS|SSS}
\toprule
    {} & {Freq.} & {FC} & {ATTN} & {HS} & {VP-50} & {VP} \\ \midrule
    {GPT2 XL} & {0.053} & {0.053} & {0.046} & {0.041} & {0.045} & {0.040} \\
    {Pythia 6B} & {0.063} & {0.052} & {0.048} & {0.045} & {0.047} & {0.042} \\
    {Pythia 12B} & {0.069} & {0.059} & {0.051} & {0.056} & {0.052} & {0.053} \\
    {LLaMA2 7B} & {0.069} & {0.068} & {0.057} & {0.051} & {0.062} & {0.053} \\
    {LLaMA2 13B} & {0.073} & {0.061} & {0.060} & {0.053} & {0.061} & {0.053} \\
    {Vicuna 13B} & {0.086} & {0.074} & {0.071} & {0.064} & {0.072} & {0.062} \\
    \bottomrule
\end{tabular}
\caption{MSE for \method{} QA Probes between predicted 
\method{} scores and QA accuracy scores.}
\label{tab:qa_accuracy_mse}
\end{table}

\subsection{Interpretability-Performance Tradeoff for \method{} VP-k}
\label{subsec:diminish_returns}
\begin{figure}[t]
  \centering
   \includegraphics[scale=.12]{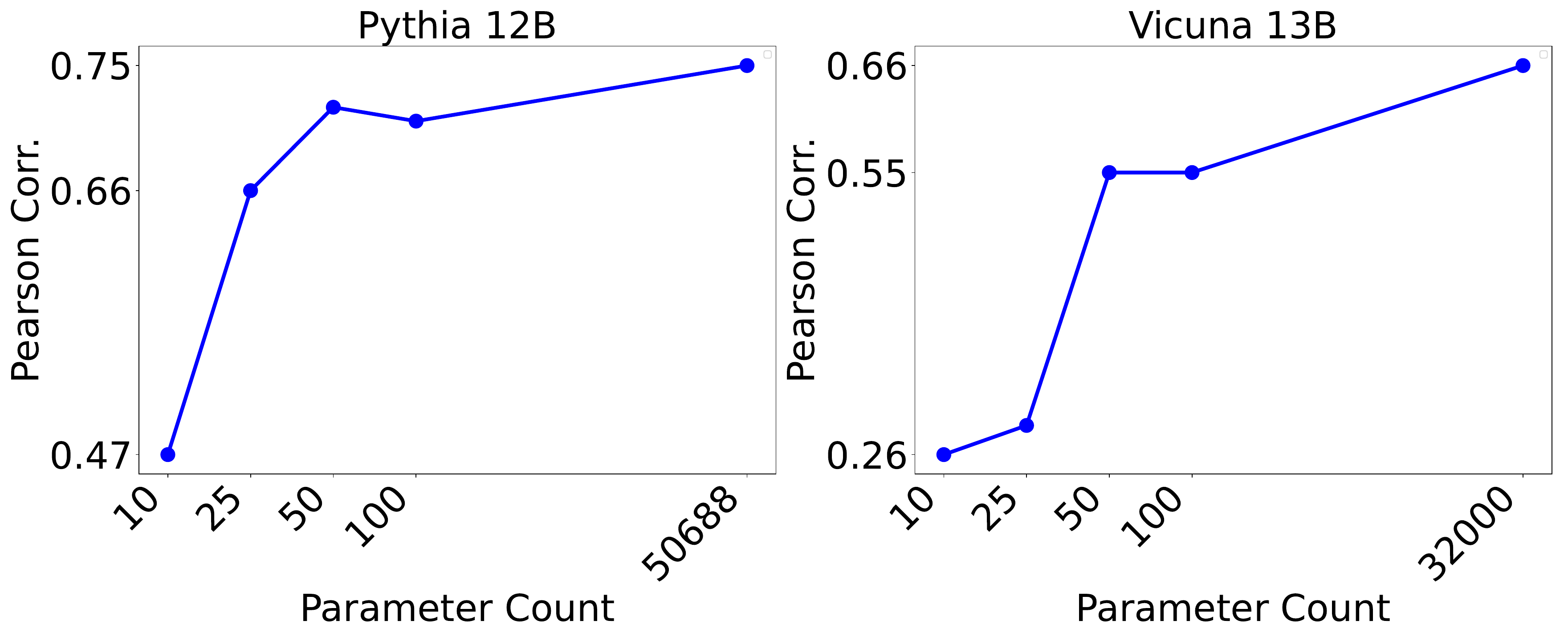}
  \caption{Correlation of \method{} OEG VP probe scores and FActScore as a function of input parameter count.}
  \label{fig:oeg_probe_main_lineplots}
\end{figure}

\begin{figure}[ht]
  \centering
  \includegraphics[scale=.12]{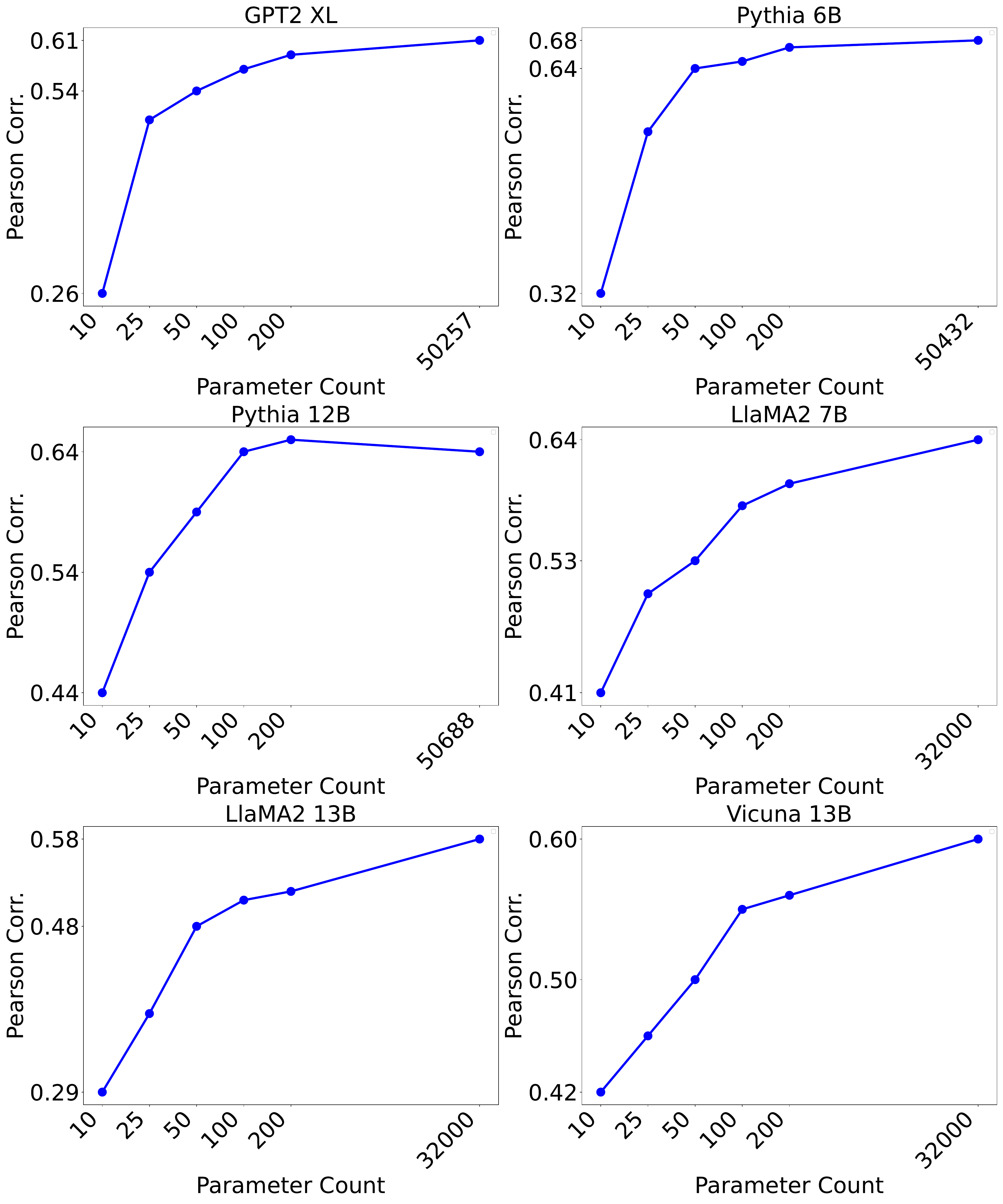}
  \caption{Correlation of \method{} QA VP probe scores and QA accuracy as a function of input parameter count.}
  \label{fig:qa_probe_vocab_proj_qa_accuracy_linegraphs}
\end{figure}

Figure~\ref{fig:oeg_probe_main_lineplots} and Figure~\ref{fig:qa_probe_vocab_proj_qa_accuracy_linegraphs} demonstrate diminishing returns in increasing the parameter count beyond 50 tokens, suggesting that a small set of tokens contains significant signals for estimating entity knowledge. There is a clear trade-off between the greater interpretability of smaller \method{} VP probes and the reduced correlation. However, the \method{} VP-50 variant remains highly interpretable with a minimal number of tokens and does not suffer a substantial correlation decline. Consequently, we chose to focus on evaluating the \method{} VP and VP-50 variants.

\subsection{Recovering QA accuracy of non-target subjects through patching}
\label{subsec:recover_qa}
The source model used in both patching experiments was LLaMA2 7B, fine-tuned for 100 epochs on passages from the Wikipedia page about \texttt{Adil Shamoo}. For both experiments, we analyzed 85 non-target subjects with $\geq 5$ questions whose QA accuracy decreased by at least 5\% and \method{} scores either increased or stayed constant post fine-tuning. Representations from upper intermediate layers (20-23) of the source model were patched into the penultimate layer (30) of the target model. In the PT layer experiment, the target model was pre-trained LLaMA2 7B, while in the FT subj. experiment, the source and target models were both the fine-tuned LLaMA2 7B model mentioned above. In computing patched accuracy, a question was marked as correct if patching from any source layer recovered the correct answer. Formally, patched QA accuracy per subject is computed according to $y_{QA,\text{patched}}^{(s)} := \frac{1}{n} \sum_{i=1}^{n} \mathbbm{1} \left[\exists \ell \in \{20, 21, 22, 23\} : \hat{a}_i^{\text{patched}(\ell)} = a_i \right]$, where $\hat{a}_i^{\text{patched}(\ell)}$ denotes the answer generated after patching from source layer $\ell$ into the target model, as described above.

While \S\ref{subsec:training} discusses patching results aggregated across all 85 non-target entities, Table~\ref{tab:recover_qa_accuracy_per_subject} presents per-entity results from both patching experiments, showing that QA accuracy can be effectively recovered from intermediate fine-tuned representations.

\begin{table*}[t]
\centering
\footnotesize
\begin{tabular}{ccccc|cc}
\toprule
    & {\method{} PT} & {\method{} FT} & {Acc. PT} & {Acc. FT} & {PT layer} & {FT subj.} \\
    \midrule 
        John, King of England & 0.51 & 0.51 & 0.67 & 0.30 & 0.50 & 0.50\\
        Maurice Le Boucher & 0.32 & 0.14 & 0.29 & 0.14 & 0.43 & 0.29\\
        The Deep & 0.51 & 0.52 & 0.71 & 0.00 & 0.71 & 0.71\\
        A Whole New World & 0.45 & 0.50 & 0.40 & 0.20 & 0.40 & 0.80\\
        Alexander the Great & 0.56 & 0.59 & 0.60 & 0.36 & 0.55 & 0.55\\
        Jaakko Laakso & 0.42 & 0.43 & 0.63 & 0.13 & 0.50 & 0.38 \\
        Reds & 0.48 & 0.49 & 0.86 & 0.29 & 0.71 & 0.71\\
        WarGames & 0.59 & 0.63 & 0.50 & 0.25 & 0.50 & 0.50\\
        Charles V & 0.33 & 0.34 & 0.54 & 0.23 & 0.77 & 0.62\\
        Recovery & 0.25 & 0.29 & 0.17 & 0.00 & 0.67 & 0.50\\
        James I of Scotland & 0.47 & 0.47 & 0.45 & 0.18 & 0.55 & 0.46\\
        Mai Van Hoa & 0.33 & 0.36 & 0.29 & 0.14 & 0.50 & 0.17\\
        The Natural & 0.48 & 0.50 & 0.63 & 0.13 & 0.50 & 0.50\\
        The Bourne Legacy & 0.67 & 0.69 & 0.67 & 0.22 & 0.67 & 0.56\\
        Beck & 0.48 & 0.51 & 0.44 & 0.22 & 0.78 & 0.56\\
        Radek Opršal & 0.46 & 0.48 & 0.57 & 0.14 & 0.43 &0.29\\
        Elena Romagnolo & 0.42 & 0.46 & 0.50 & 0.17 & 0.33 & 0.33\\
        Jaroslav Kocián & 0.47 & 0.56 & 0.38 & 0.13 & 0.50 & 0.25\\
    \bottomrule
\end{tabular}
\caption{Per-entity results from the patching experiments discussed in \S\ref{subsec:training}. The recovery of QA accuracy is evident from both the observed increase in accuracy over Acc. FT and similarity to Acc. PT after patching (PT layer, FT subj.). Also, the alignment of \method{} PT and FT scores with both Acc. PT and patched accuracy indicates that \method{} reflects the amount of information encoded in the representations rather than in the observed output of the fine-tuned model.}
\label{tab:recover_qa_accuracy_per_subject}
\end{table*}

\subsection{Statistical significance of correlations}
Table~\ref{tab:p_vals_qa}, Table~\ref{tab:p_vals_oeg}, Table~\ref{tab:p_vals_transfer} represent the p-values associated with the Pearson correlation values found in Table~\ref{tab:qa_accuracy_corr}, Table~\ref{tab:factscore_corr_oeg}, Table~\ref{tab:factscore_corr_qa}, respectively.

\begin{table*}[t]
\centering
\begin{tabular}{lcccccc}
\toprule
{Model} & {Freq.} & {FC} & {ATTN} & {VP-50} & {VP} & {HS} \\ \midrule
GPT2 XL & ${2.54e^{-17}}$ & ${8.24e^{-47}}$ & ${7.70e^{-56}}$ & ${1.72e^{-59}}$ & ${9.80e^{-79}}$ & ${1.73e^{-75}}$ \\
Pythia 6B & ${3.26e^{-19}}$ & ${6.41e^{-75}}$ & ${2.74e^{-85}}$ & ${7.28e^{-90}}$ & ${1.33e^{-106}}$ & ${3.16e^{-106}}$ \\
Pythia 12B & ${1.03e^{-15}}$ & ${4.46e^{-62}}$ & ${2.25e^{-76}}$ & ${2.53e^{-74}}$ & ${8.72e^{-91}}$ & ${6.46e^{-91}}$ \\
LLaMA2 7B & ${9.96e^{-15}}$ & ${4.54e^{-49}}$ & ${1.17e^{-70}}$ & ${3.07e^{-56}}$ & ${4.50e^{-89}}$ & ${2.52e^{-89}}$\\
LLaMA2 13B & ${7.81e^{-13}}$ & ${6.77e^{-49}}$ & ${1.50e^{-50}}$ & ${1.98e^{-45}}$ & ${3.43e^{-70}}$ & ${1.74e^{-71}}$ \\
Vicuna 13B & ${5.15e^{-13}}$ & ${1.29e^{-48}}$ & ${1.68e^{-53}}$ & ${3.00e^{-50}}$ & ${2.07e^{-77}}$ & ${3.25e^{-76}}$ \\
\bottomrule
\end{tabular}
\caption{p-values for different models across \method{} QA probes.}
\label{tab:p_vals_qa}
\end{table*}

\begin{table*}[t]
\centering
\begin{tabular}{lcccccc}
\toprule
{Model} & {Freq.} & {FC} & {ATTN} & {VP-50} & {VP} & {HS} \\ \midrule
Pythia 12B & ${1.00e^{-2}}$ & ${2.68e^{-6}}$ & ${7.20e^{-11}}$ & ${4.57e^{-9}}$ & ${3.04e^{-10}}$ & ${4.21e^{-11}}$ \\
Vicuna 13B & $2.05e^{-2}$ & ${1.28e^{-3}}$ & ${6.16e^{-6}}$ & ${2.50e^{-4}}$ & ${3.43e^{-6}}$ & ${4.02e^{-6}}$ \\
\bottomrule
\end{tabular}
\caption{p-values for different models across \method{} OEG probes}
\label{tab:p_vals_oeg}
\end{table*}

\begin{table*}[t]
\centering
\begin{tabular}{lcccccc}
\toprule
{Model} & {Freq.} & {FC} & {ATTN} & {VP-50} & {VP} & {HS} \\ \midrule
Pythia 12B & ${2.74e^{-29}}$ & ${1.00e^{-6}}$ & ${8.43e^{-7}}$ & ${1.76e^{-2}}$ & ${4.92e^{-5}}$ & ${3.25e^{-7}}$ \\
Vicuna 13B & ${2.28e^{-16}}$ & ${4.80e^{-4}}$ & ${1.39e^{-4}}$ & ${7.84e^{-5}}$ & ${9.18e^{-6}}$ & ${2.12e^{-5}}$ \\
\bottomrule
\end{tabular}
\caption{p-values for the correlation of \method{} QA probe and FActScore scores.}
\label{tab:p_vals_transfer}
\end{table*}

\subsection{QA correlation plots}
\label{subsec:qa_corr_plots_more}
\begin{figure*}[ht]
  \centering
  \includegraphics[scale=.25]{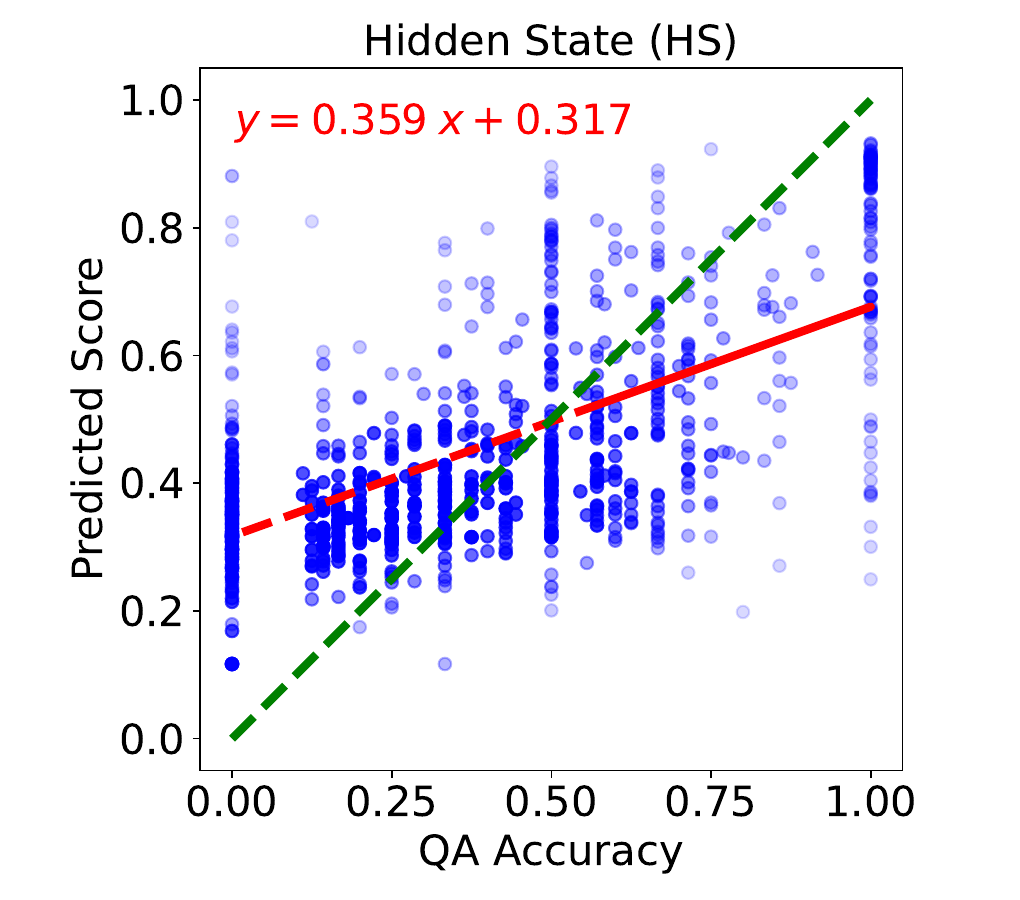}
  \includegraphics[scale=.25]{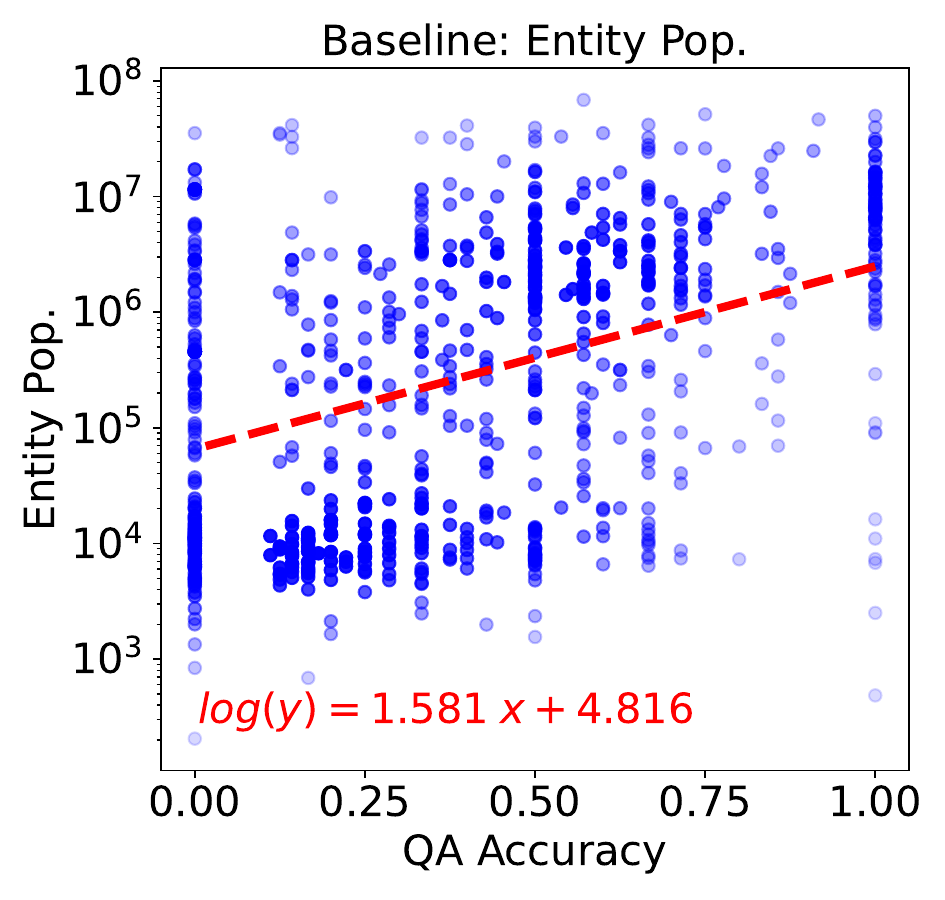}
  \includegraphics[scale=.25]{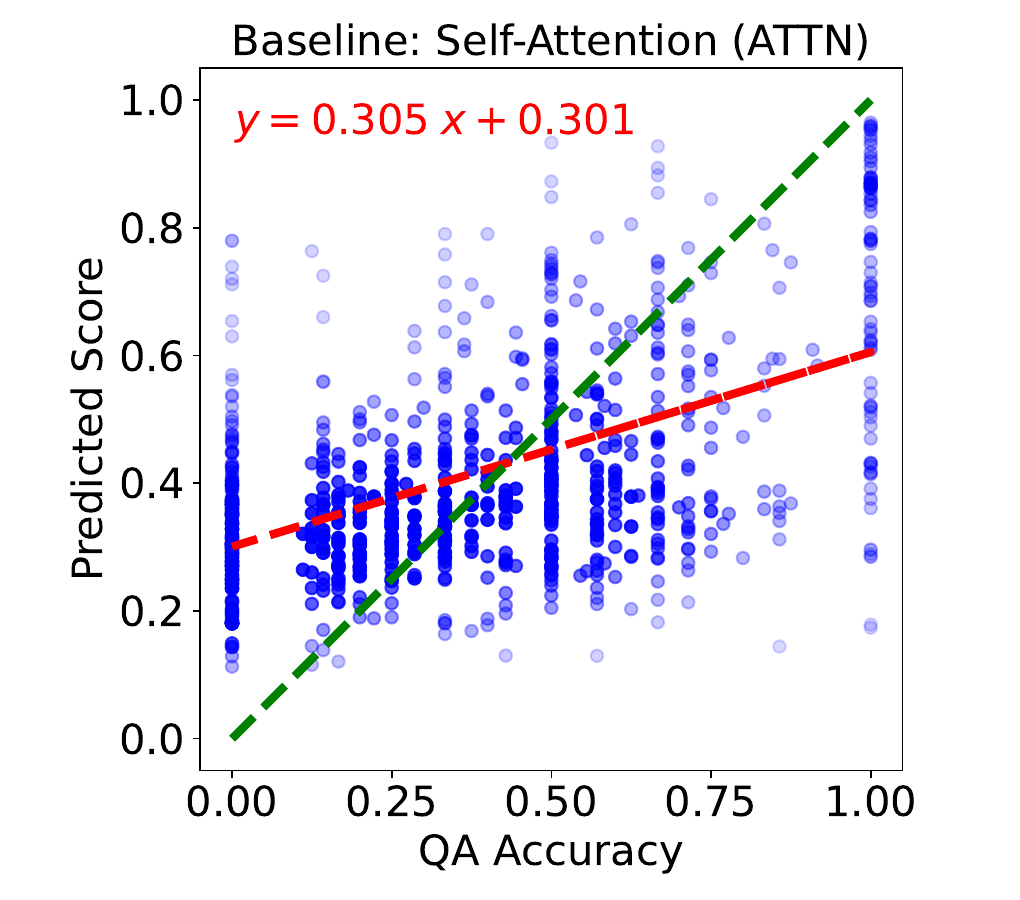}
  \includegraphics[scale=.25]{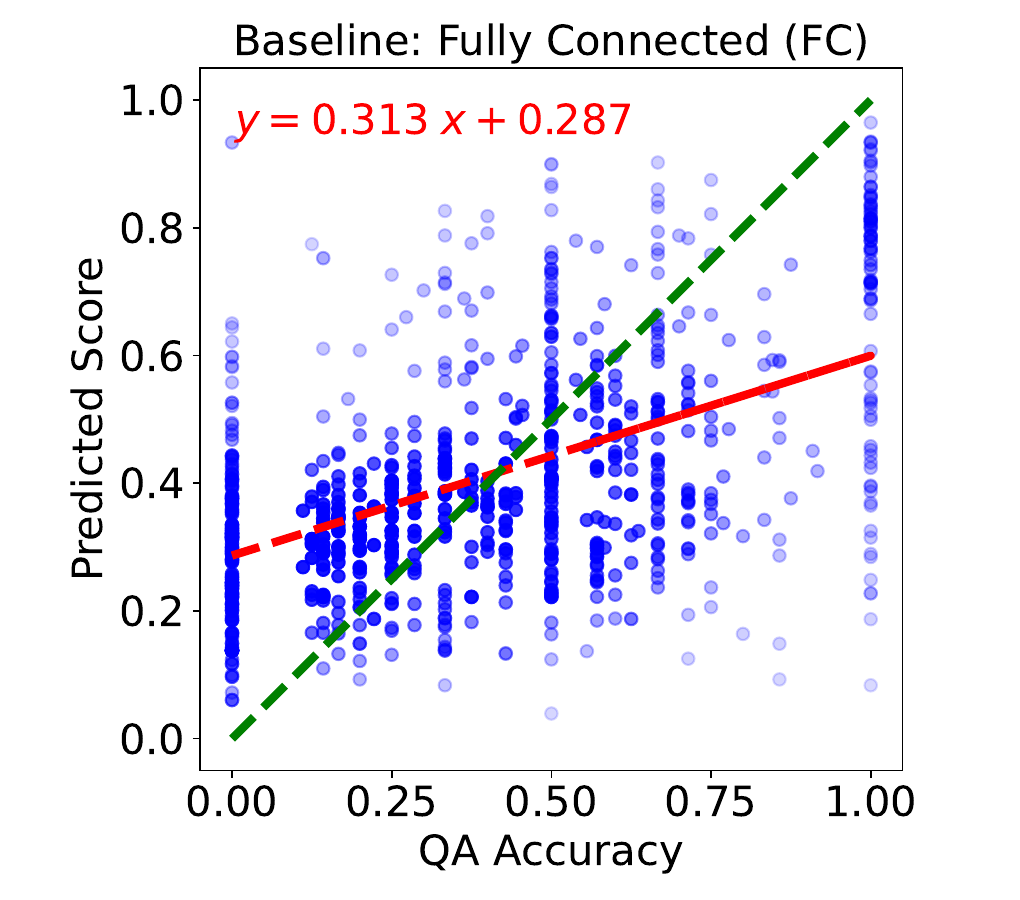}
  \caption{Vicuna 13B: Predicted scores from the \method{} QA probe versus the golden QA accuracy scores.}
  \label{fig:vicuna_13B_qa_probes_corr_qa_accuracy}
\end{figure*}

\begin{figure*}[ht]
  \centering
  \includegraphics[scale=.25]{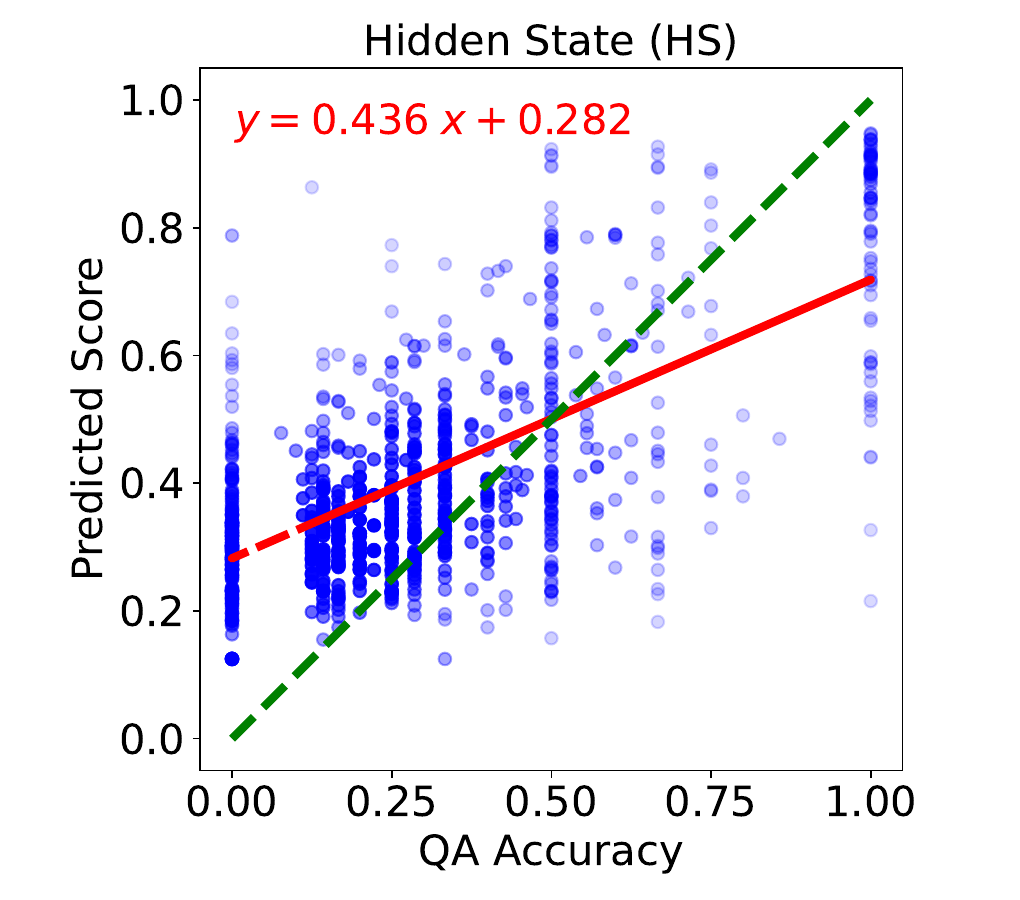}
  \includegraphics[scale=.25]{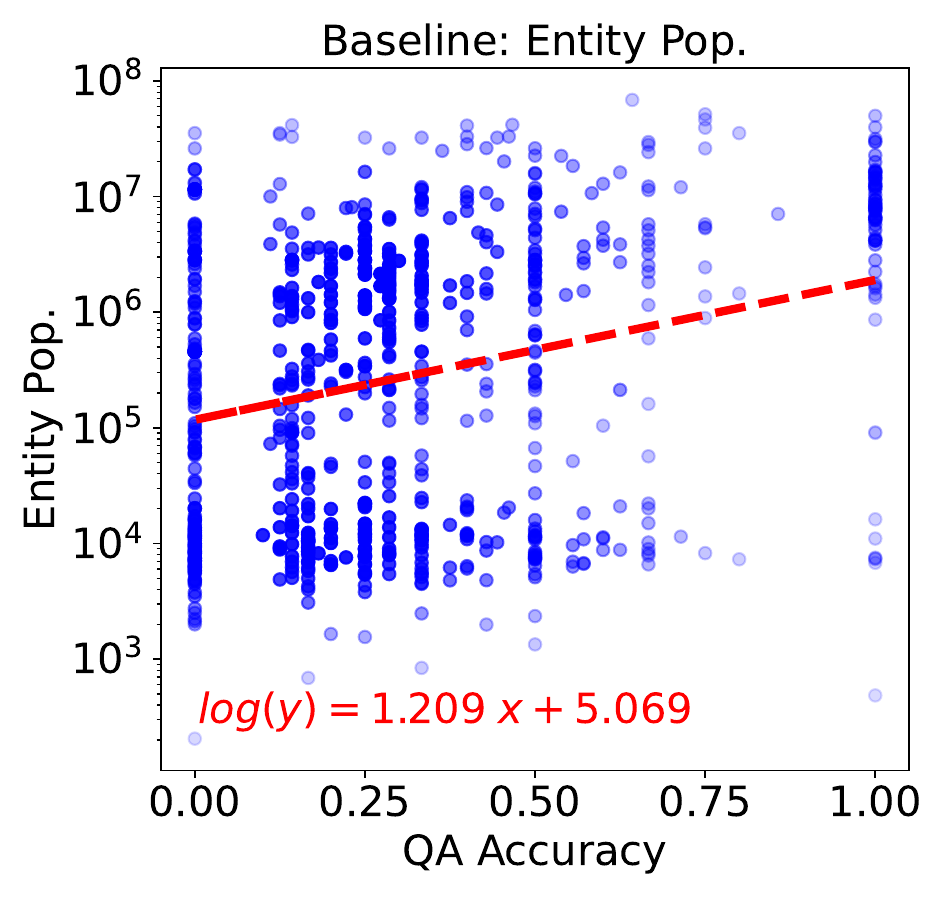}
  \includegraphics[scale=.25]{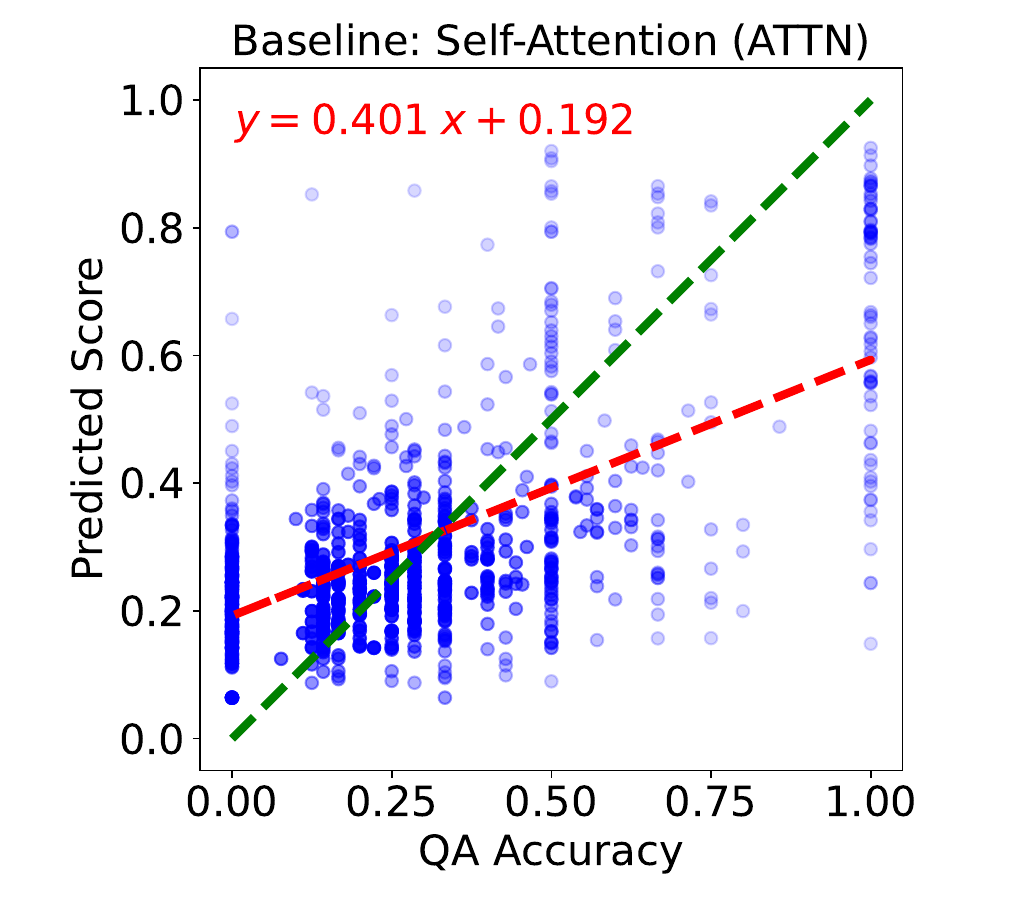}
  \includegraphics[scale=.25]{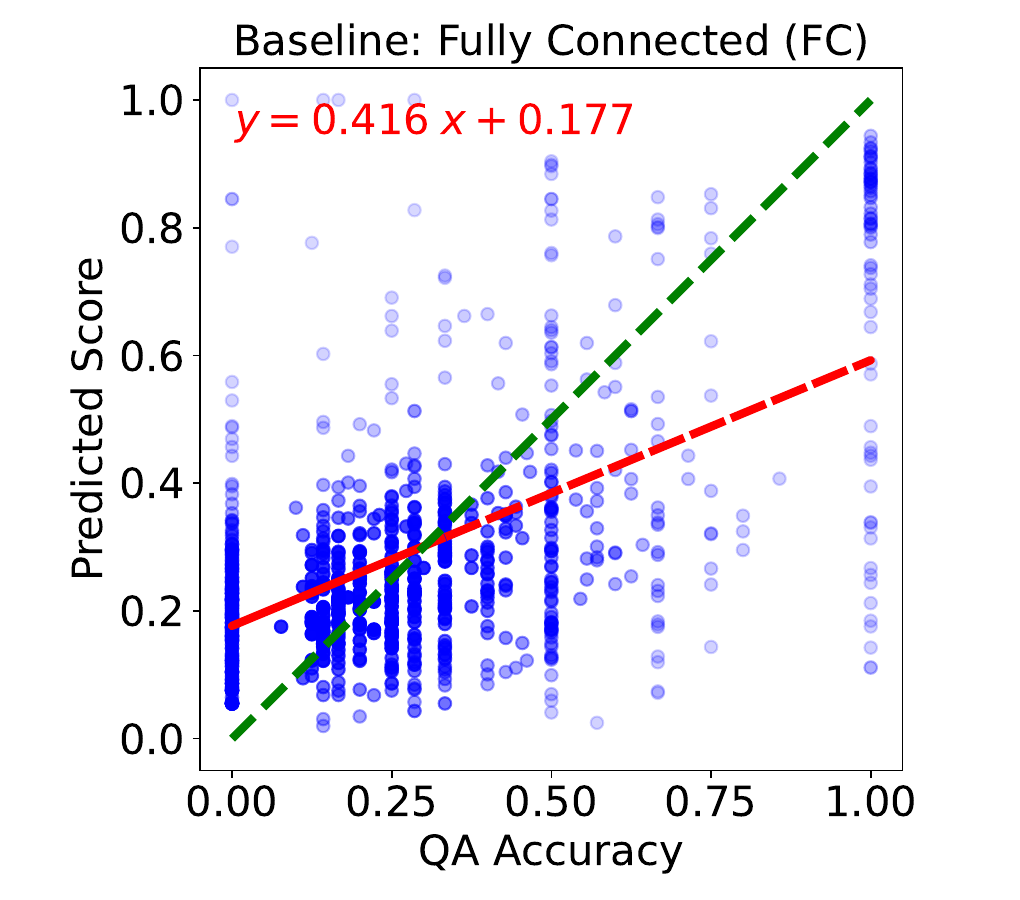}
  \caption{Pythia 12B: Predicted scores from the \method{} QA probe versus the golden QA accuracy scores.}
  \label{fig:pythia_12B_qa_probes_corr_qa_accuracy}
\end{figure*}

Figure~\ref{fig:vicuna_13B_qa_probes_corr_qa_accuracy} and Figure~\ref{fig:pythia_12B_qa_probes_corr_qa_accuracy}, show the results for the QA experiments in \S\ref{subsec:results} for Vicuna 13B and Pythia 12B, respectively.

\subsection{OEG correlation plots}
\label{subsec:oeg_corr_plots_more}

\begin{figure*}[ht]
  \centering
  \includegraphics[scale=.25]{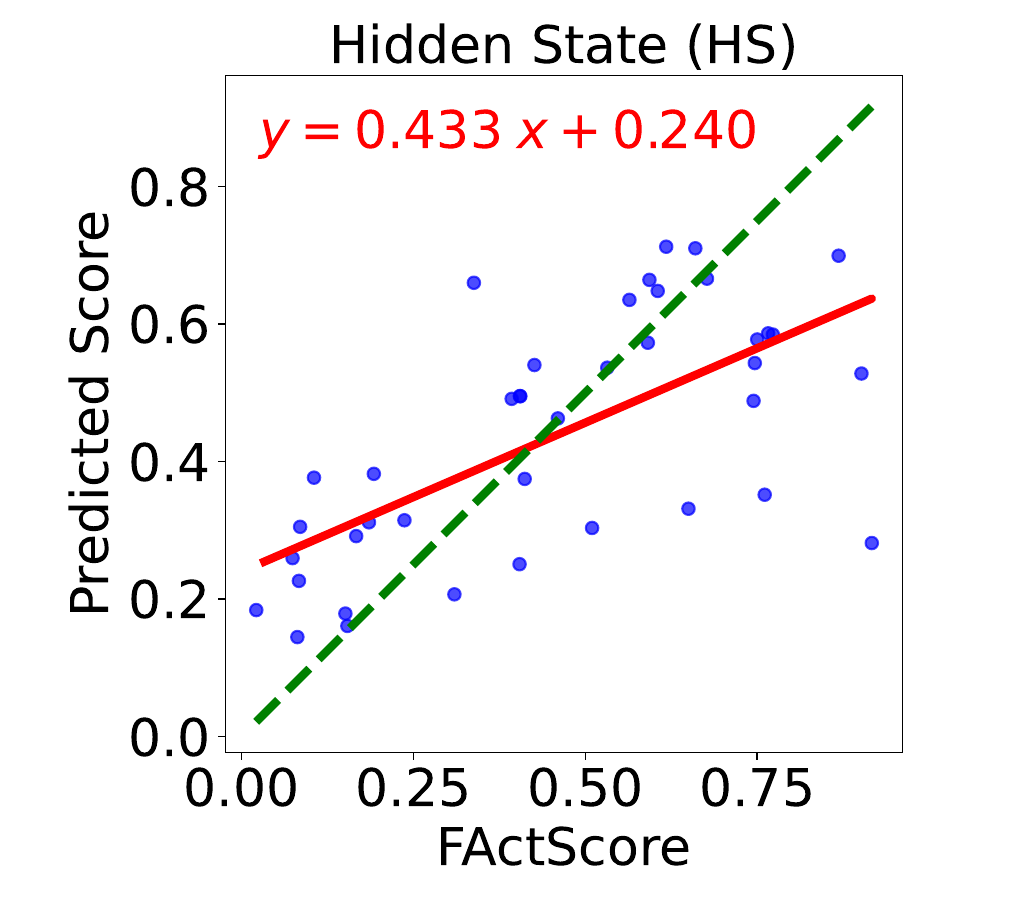}
  \includegraphics[scale=.25]{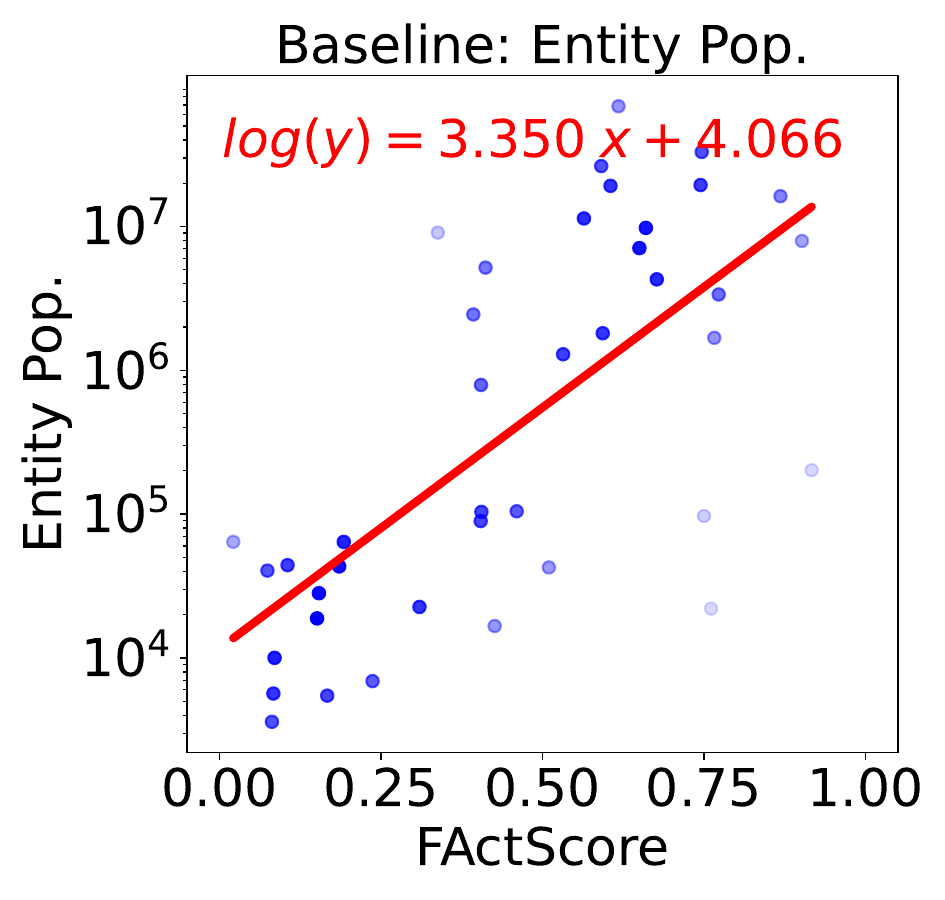}
  \includegraphics[scale=.25]{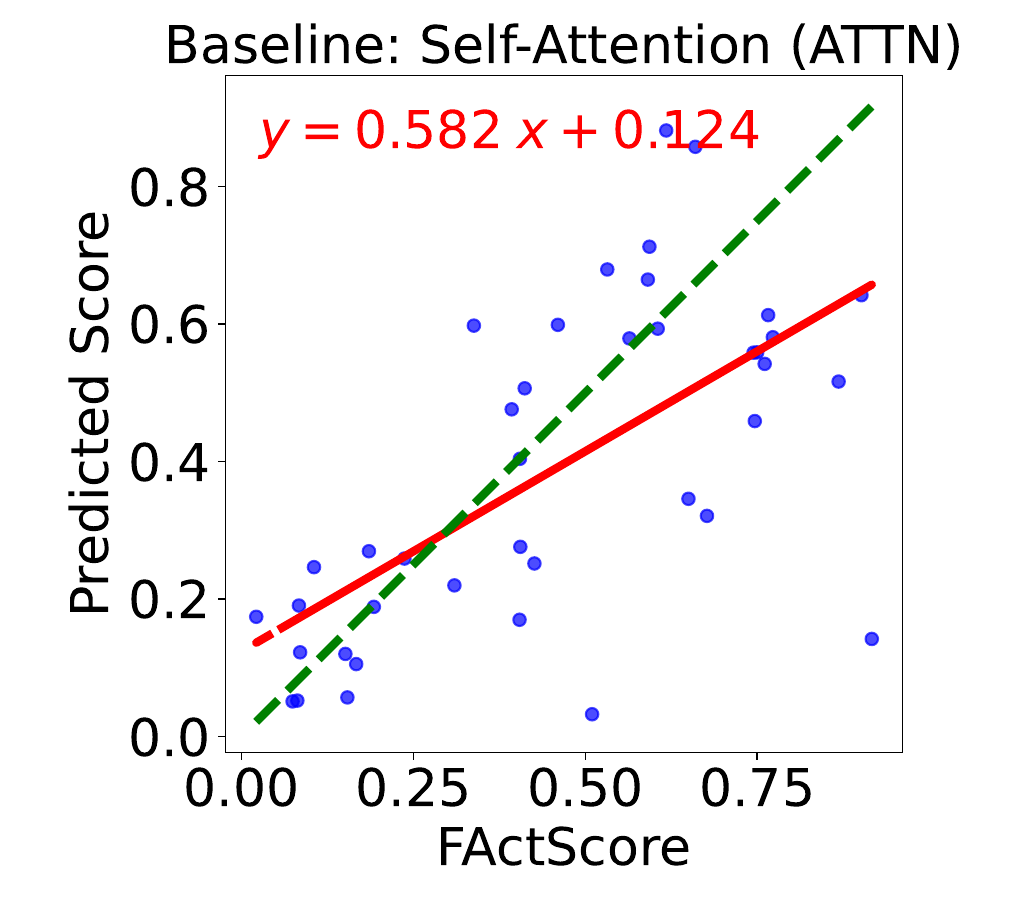}
  \includegraphics[scale=.25]{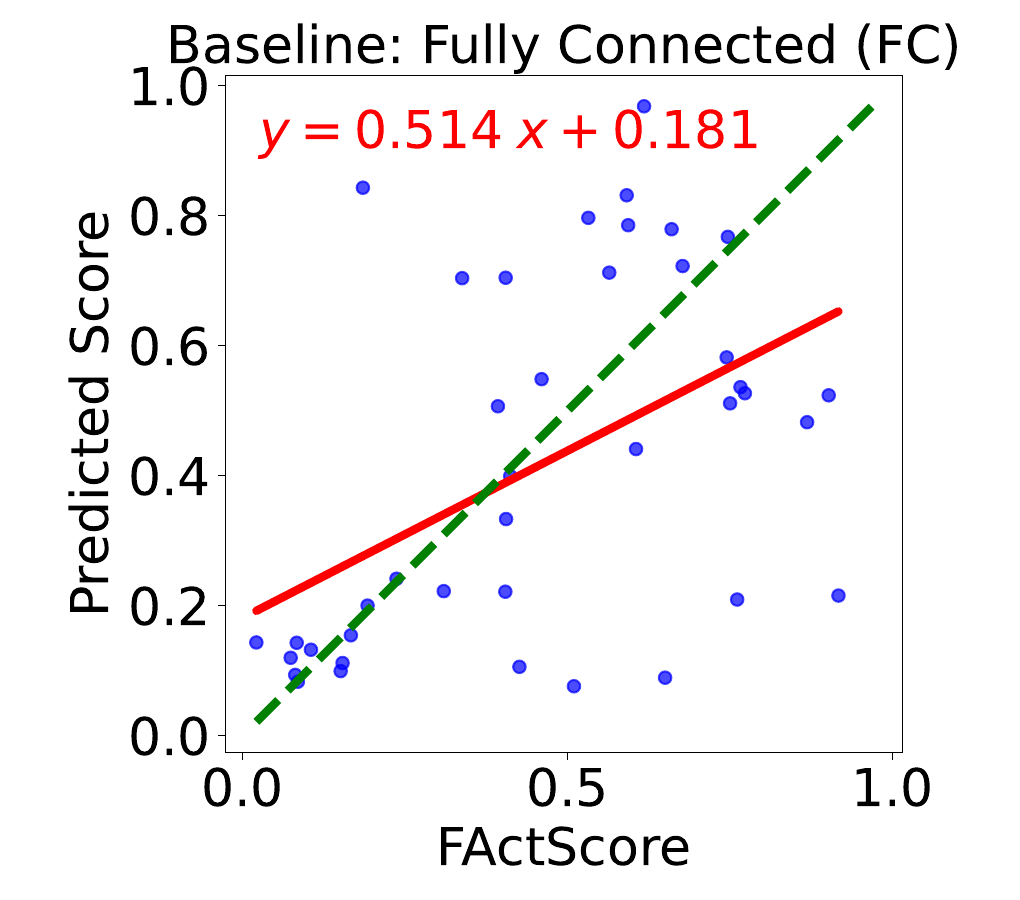}
  \caption{Vicuna 13B: Predicted scores from the \method{} OEG probe versus the golden FActScore scores.}
  \label{fig:vicuna_13B_factscore_probes_corr_factscore}
\end{figure*}

\begin{figure*}[ht]
  \centering
  \includegraphics[scale=.25]{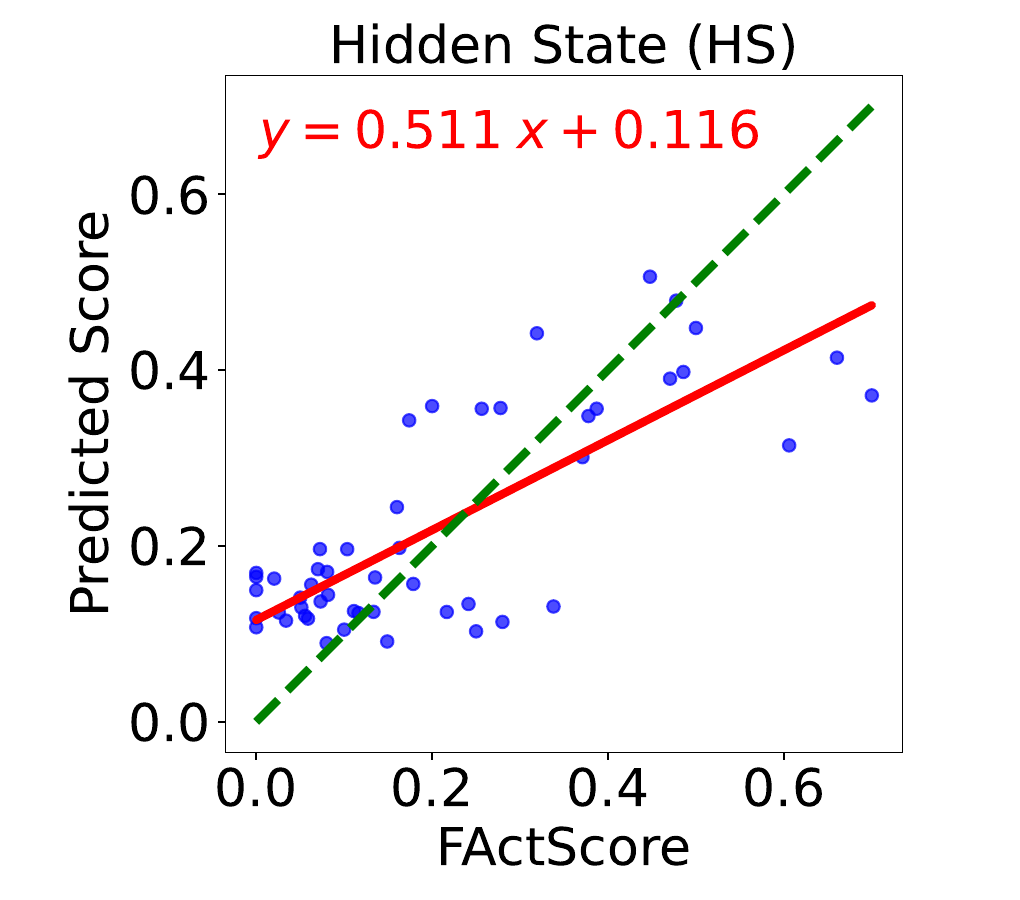}
  \includegraphics[scale=.25]{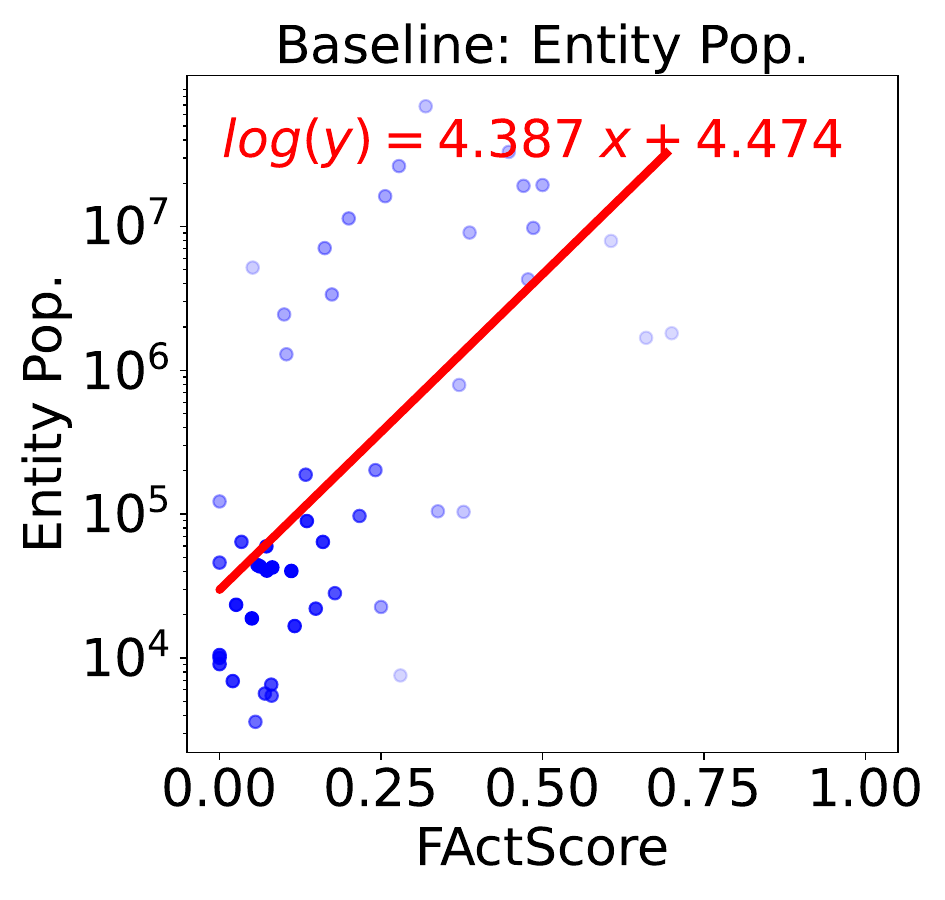}
  \includegraphics[scale=.25]{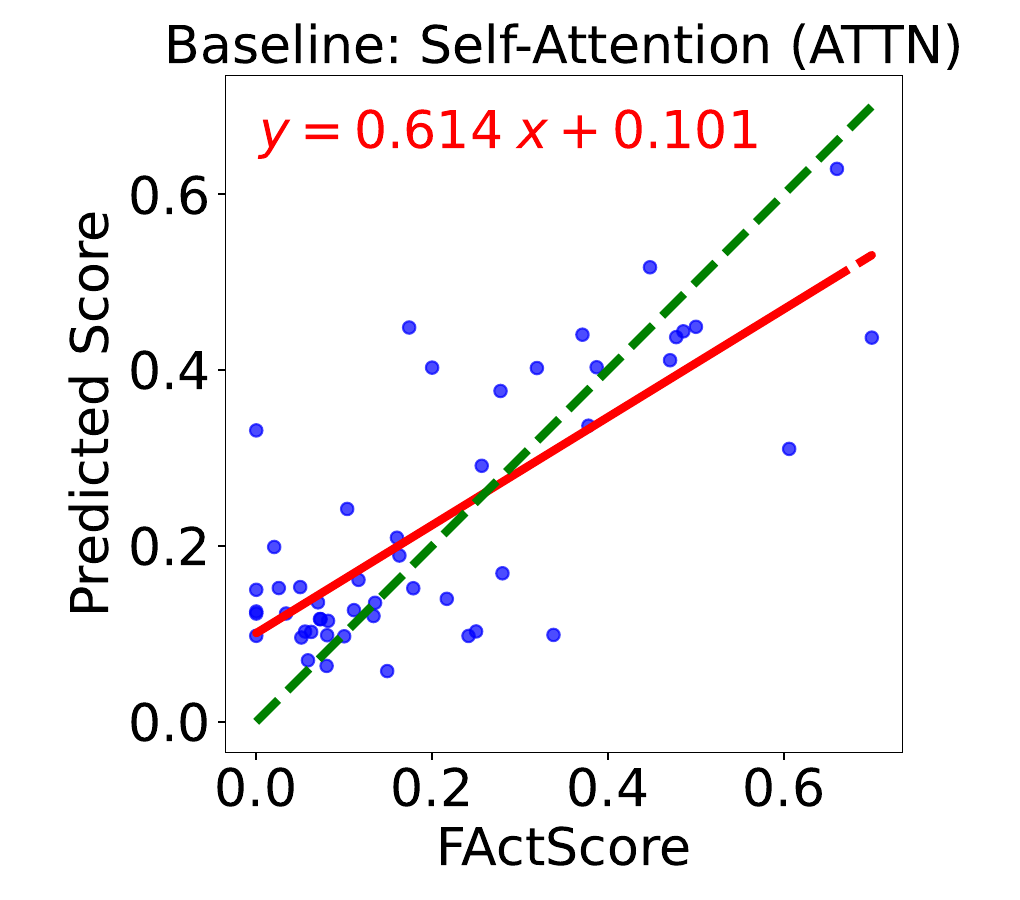}
  \includegraphics[scale=.25]{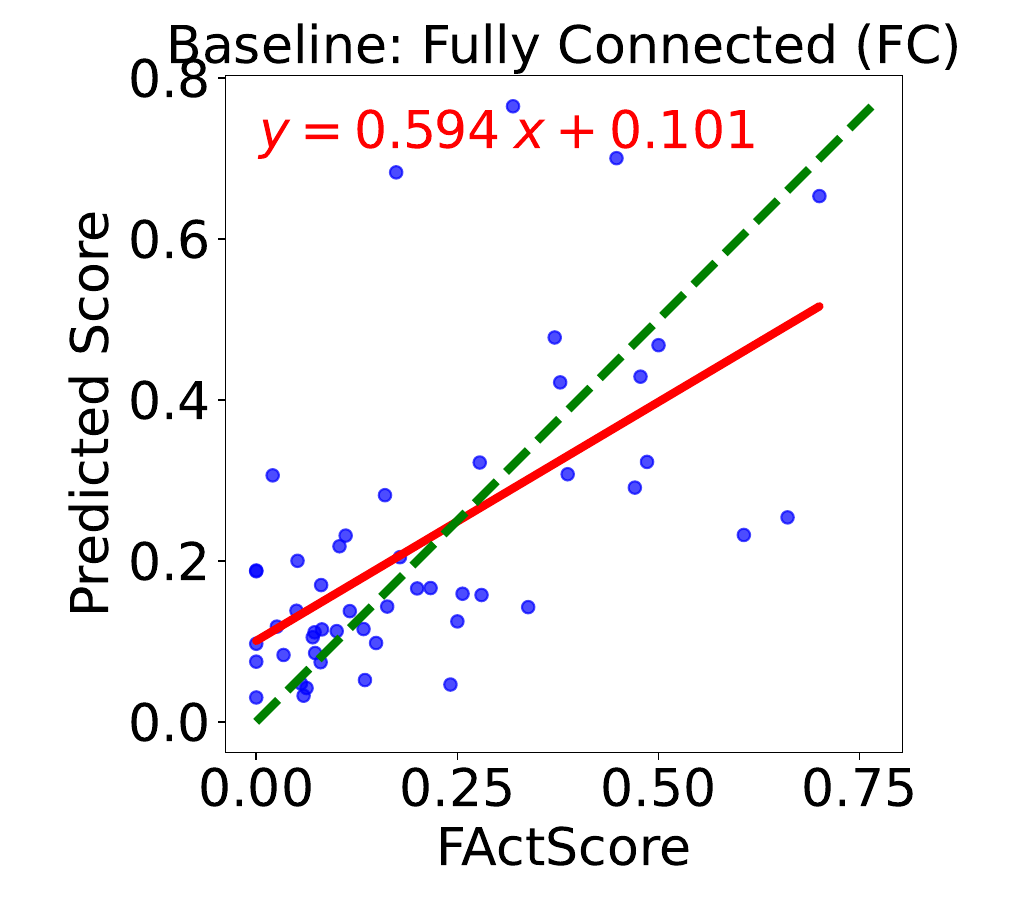}
  \caption{Pythia 12B: Predicted scores from the \method{} OEG probe versus the golden FActScore scores.}
  \label{fig:pythia_12B_factscore_probes_corr_factscore}
\end{figure*}

Figure~\ref{fig:vicuna_13B_factscore_probes_corr_factscore} and  Figure~\ref{fig:pythia_12B_factscore_probes_corr_factscore} show results for the OEG experiments in \S\ref{subsec:results}.

\end{document}